\documentclass[10pt,twocolumn,letterpaper]{article}

\usepackage{cvpr}             

\usepackage{booktabs}
\usepackage{graphicx}

\definecolor{cvprblue}{rgb}{0.21,0.49,0.74}
\usepackage[pagebackref,breaklinks,colorlinks,allcolors=cvprblue]{hyperref}

\def\paperID{*****} 
\def\confName{CVPR}
\def\confYear{2026}

\title{Benchmarking stereo reconstruction for 3D printable Martian terrain models}

\author{Josephine Wang\\
MIT\\
Cambridge, MA, USA\\
{\tt\small josiexw@mit.edu}
}

\begin{document}
\maketitle
\begin{abstract}
Reconstructing printable 3D models from Mars rover imagery is challenging because Martian terrain is low-texture, irregular, and partially observed. We evaluate a pipeline that estimates stereo depth from NASA Curiosity images, completes geometry, and exports watertight OBJ meshes. On Middlebury, RAFT-Stereo outperforms semi-global block matching (SGBM), reducing disparity MAE from $3.22\,$px to $0.73\,$px and increasing valid prediction coverage from $76.3$\% to $100.0$\%. On Curiosity imagery, however, RAFT’s denser disparities show weaker edge alignment and higher photometric reprojection error, suggesting that benchmark accuracy does not directly transfer to Martian terrain reconstruction. Geometry completion demonstrates a tradeoff between local fidelity and global connectivity. We find that alpha shapes preserve accurate but fragmented structure, Poisson reconstruction produces more coherent meshes but adds unsupported surfaces, and a deterministic diffusion-fill baseline is intermediate but sensitive to stereo quality. Overall, standard stereo and completion methods can produce printable approximations of Martian terrain, but reliable reconstruction requires stronger domain-specific validation.
\end{abstract}
    
\section{Introduction}

NASA's Curiosity rover has collected stereo photographs of Martian terrain since 2012. These images provide a detailed record of the planet's surface, but converting them into accurate 3D models remains difficult due to low-texture rocks, uneven boundaries, variable illumination, and incomplete views. Reliable 3D reconstruction from rover imagery would make Martian terrain easier to analyze and reproduce as tactile models. We study a pipeline for estimating depth from Curiosity stereo imagery, completing terrain geometry, and exporting printable 3D models.

\paragraph{Related work.}
Prior work in 3D reconstruction often combines geometric constraints with learned shape priors. Kar et al.'s multi-view stereo machine follows this approach by using known camera poses to unproject image features into a shared 3D volume, fuse them across views, and reason over them with a learned network \cite{kar2017}. However, Kar et al. evaluate on clean, object-centered ShapeNet renderings with known poses and voxelized ground-truth, whereas our setting uses rover imagery without ground-truth geometry \cite{shapenet}.

For stereo reconstruction, we compare SGBM, a classical semi-global matching method that has been used in planetary settings, with RAFT-Stereo, a learned recurrent stereo method that performs well on standard benchmarks \cite{sgm2008,raftstereo}. We also evaluate methods for converting partial point clouds into printable meshes, comparing adaptive alpha shapes, Poisson surface reconstruction, and a deterministic diffusion-fill baseline \cite{edelsbrunner1994threedalpha, kazhdan2006poisson, 3dgan}.

\paragraph{Contributions.}
We make the following contributions. (i) \textbf{Domain-shift analysis for planetary stereo reconstruction.} We compare SGBM and RAFT-Stereo on Curiosity imagery and evaluate how standard benchmark performance transfers to Mars terrain \cite{sgm2008,raftstereo}. (ii) \textbf{Occluded surface estimation.} We compare adaptive alpha shapes, Poisson reconstruction, and a deterministic diffusion-fill baseline for converting partial point clouds into completed meshes. (iii) \textbf{Printable mesh evaluation.} We produce watertight printable OBJ files through repairing reconstructed geometry. (iv) \textbf{Planetary benchmarking.} We evaluate the gap between standard stereo benchmarks and Curiosity imagery using Middlebury ground truth and rover-image proxy metrics. Overall, this study shows both the limitations and potential of stereo reconstruction for printable Martian terrain models.

\section{Methods}

\subsection{Data collection}

We constructed a stereo dataset from NASA Mars Science Laboratory Curiosity imagery \cite{mslrawimages}. We queried the raw image archive for navigation-camera images, matched left-right pairs using spacecraft clock timestamps, and resized each image to $640\times480$ before stereo reconstruction. The final dataset contained 100 left-right image pairs.

To separate standard benchmark performance from target-domain behavior, we evaluated stereo reconstruction on both Middlebury stereo scenes and Curiosity imagery. Middlebury provides calibrated stereo pairs with ground-truth disparities and masks for direct quantitative error evaluation \cite{scharstein2002taxonomy}. Curiosity imagery does not provide these ground-truth disparities, so we evaluated it using reprojection, edge-alignment, disparity/depth, and downstream mesh-quality metrics.

\subsection{Stereo reconstruction}

Given a stereo pair $(I_L, I_R)$, each method predicted a disparity map $d(u,v)$ in pixels. We converted valid disparities to metric depth using
\begin{equation}
    z(u,v) = \frac{f_x B}{d(u,v)},
\end{equation}
where $f_x$ is the focal length in pixels and $B$ is the stereo baseline. For each Curiosity pair, we loaded camera parameters from the corresponding calibration file, resized images to the calibrated resolution, and discarded non-finite, non-positive, and physically implausible depths before point-cloud reconstruction.

\subsubsection{Semi-global block matching}

We used OpenCV's semi-global block matching (SGBM) function as a classical stereo baseline, following the SGM algorithm of Hirschm{\"u}ller \cite{sgm2008}. We used a block size of 7, 96 disparity levels, a uniqueness ratio of 10, speckle filtering, three-way SGBM mode, and weighted least-squares disparity filtering with the corresponding right-view matcher.

\subsubsection{RAFT-Stereo}

As a learned stereo baseline, we used RAFT-Stereo, a recurrent architecture for rectified stereo matching \cite{raftstereo}. We used the Middlebury-trained checkpoint, ran 32 update iterations, converted grayscale inputs to three-channel tensors, padded them to multiples of 32, and negated the horizontal optical flow output to obtain disparity.

\subsubsection{Evaluation}

For Curiosity pairs, we evaluated each disparity map using reprojection, edge-alignment, disparity/depth, and downstream mesh-quality metrics (Appendix).

\subsubsection{Benchmarking}

We benchmarked both methods on Middlebury and Curiosity imagery. Middlebury provides ground-truth disparity, so we reported valid prediction ratio, MAE, RMSE, and bad-pixel rates \cite{scharstein2002taxonomy}. Curiosity does not provide dense ground truth, so we used the metrics above.

\subsection{Occlusion estimation}

To estimate missing geometry, we converted each valid depth map into a 3D point cloud by back-projecting pixels through the camera model:
\begin{equation}
    x = \frac{(u-c_x)z}{f_x}, \qquad
    y = -\frac{(v-c_y)z}{f_y}, \qquad
    z = \frac{f_xB}{d}.
\end{equation}
We subsampled the depth map with stride 2, rejected depths outside $[0.05,10.0]$\,m, voxel-downsampled the point cloud, removed outliers, and capped the point count for stable reconstruction.

To isolate the contribution of each completion method, we evaluated the following reconstruction configurations: alpha shapes alone, Poisson reconstruction alone, and a hybrid that combines both. Alpha shapes test local boundary preservation, Poisson reconstruction tests global connectivity, and the hybrid tests whether concatenating the two outputs before repair recovers the strengths of each \cite{edelsbrunner1983alpha, kazhdan2006poisson}. We additionally compared against a deterministic diffusion-fill baseline with fixed height perturbations. Although we cite learned 3D completion work for context \cite{3dgan}, this baseline is not a generative model and uses no learned components.

\subsubsection{Adaptive tracing alpha shapes}

Alpha shapes reconstruct a surface by selecting a subset of the Delaunay triangulation controlled by a scale parameter $\alpha$ \cite{edelsbrunner1983alpha, edelsbrunner1994threedalpha}. We set $\alpha$ proportional to the cleaned point-cloud bounding-box diagonal:
\begin{equation}
    \alpha = \lambda_{\alpha}\,\mathrm{diag}(\mathcal{B}),
\end{equation}
where $\lambda_{\alpha}=0.02$ and $\mathcal{B}$ is the axis-aligned bounding box of the visible point cloud. This made the reconstruction scale with scene size rather than rely on a fixed world-space parameter. The resulting mesh was cropped to an expanded input bounding box and cleaned.

\subsubsection{Poisson surface reconstruction}

Poisson surface reconstruction estimates an implicit surface from oriented points by solving a spatial Poisson problem over the input normal field \cite{kazhdan2006poisson}. Before reconstruction, we estimated normals using a radius proportional to the point-cloud bounding-box diagonal and oriented them toward the camera. We ran reconstruction with octree depth $7$ and scale $1.2$, removed vertices below the 2nd percentile of the Poisson density estimate, and then cropped and cleaned the mesh.

\subsubsection{Hybrid alpha-Poisson reconstruction}

The hybrid configuration concatenates the alpha-shape and Poisson meshes before the repair stage to test whether combining local boundary fidelity with global connectivity produces better printable output than either method alone.

\subsubsection{Deterministic diffusion-fill baseline}

We implemented a baseline for filling gaps in the partial depth map. The method converts the stereo depth map into a partial depth grid, completes missing cells through iterative diffusion while keeping observed cells anchored to the measured stereo depths, adds a small fixed multi-scale height perturbation only in unobserved regions, and back-projects the completed depth grid into a triangle mesh. Finally, triangles spanning large depth discontinuities are removed.

\subsubsection{Evaluation}

For each mesh, we sampled points uniformly from the reconstructed surface and computed bidirectional nearest-neighbor distances between the visible point cloud and the mesh samples \cite{fan2017pointset}. We reported normalized Chamfer distance, visible coverage, mesh novelty, and topology statistics to distinguish dense mesh generation from reconstructions that preserved the observed surface while plausibly completing unobserved geometry (Appendix).

\subsection{Printable 3D mesh}

Finally, we converted a representative subset of the completed geometry into printable meshes for qualitative inspection. Specifically, we exported printable OBJ files for the first five SGBM scenes and first five RAFT scenes available for each completion strategy. For each exported case, we loaded the corresponding alpha-shape, Poisson, hybrid alpha-Poisson, and deterministic diffusion-fill reconstructions. We repaired each mesh using Trimesh \cite{trimesh}, voxelized it at a pitch determined by the longest side of the bounding box divided by a target resolution of $220$, extracted a watertight surface using marching cubes \cite{lorensen1987marching}, retained the largest connected component, and exported it in OBJ format.

For qualitative comparison, we rendered saved OBJ files from a fixed viewpoint. We measured printable mesh vertex count, face count, connected components, largest-component fraction, and watertightness (Appendix).

\subsection{Completion ablation}

To disentangle completion behavior from upstream stereo error, we ran a synthetic-ground-truth ablation using synthetic depth surfaces with known geometry. We generated four deterministic terrain-like height fields spanning ridges, depressions, ripples, and step discontinuities at the same image resolution as the Curiosity inputs. For each surface, we applied three structured occlusion masks: a centered hole, a diagonal shadow region, and a missing right-hand band. We then reran alpha shape, Poisson, and deterministic diffusion-fill completion on the resulting partial depth maps.

For evaluation, we converted each synthetic depth surface into a reference mesh, sampled points uniformly from that surface, and compared the completed mesh against the ground-truth samples using the same normalized Chamfer distance, coverage, novelty, and topology statistics as in the real-data evaluation. This isolates completion accuracy under missing-data patterns without stereo estimation error.

\begin{figure}[htbp!]
    \centering
    \includegraphics[width=\linewidth]{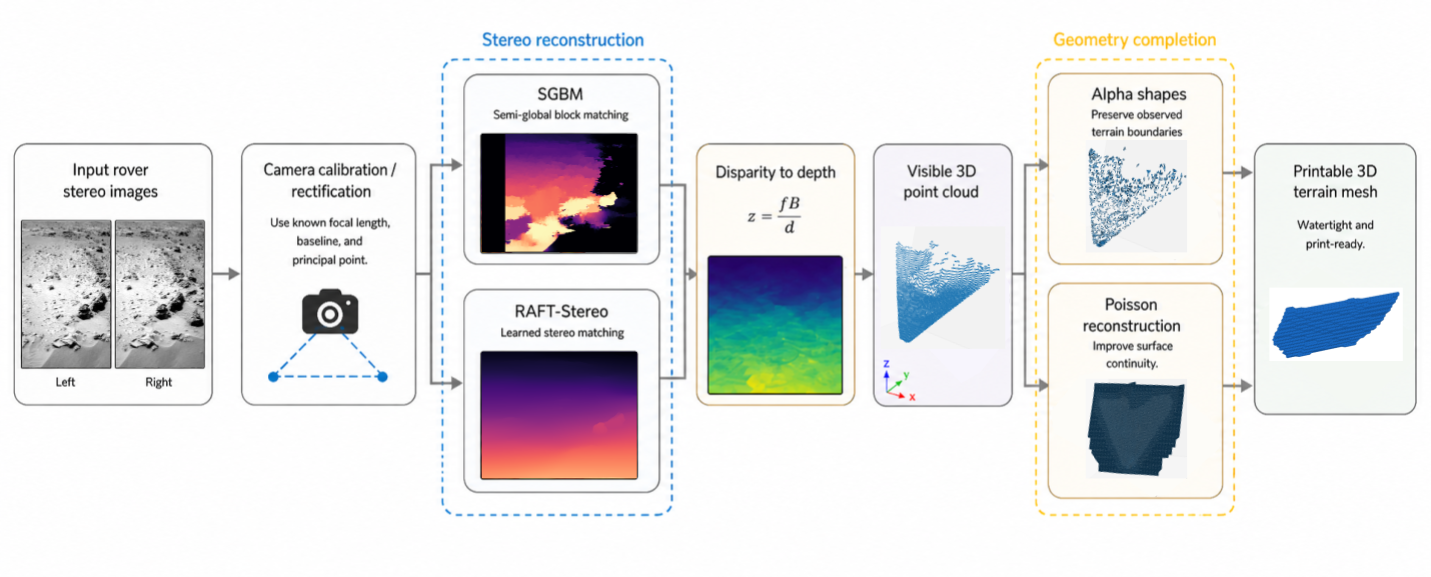}
    \caption{Overview of the pipeline for reconstructing printable 3D Martian terrain meshes from Curiosity stereo imagery.}
    \label{fig:pipeline}
\end{figure}

\section{Results}

\subsection{Stereo reconstruction: Benchmarking Middlebury vs. Curiosity datasets}

\begin{figure}[htbp]
    \centering
    \includegraphics[width=\linewidth]{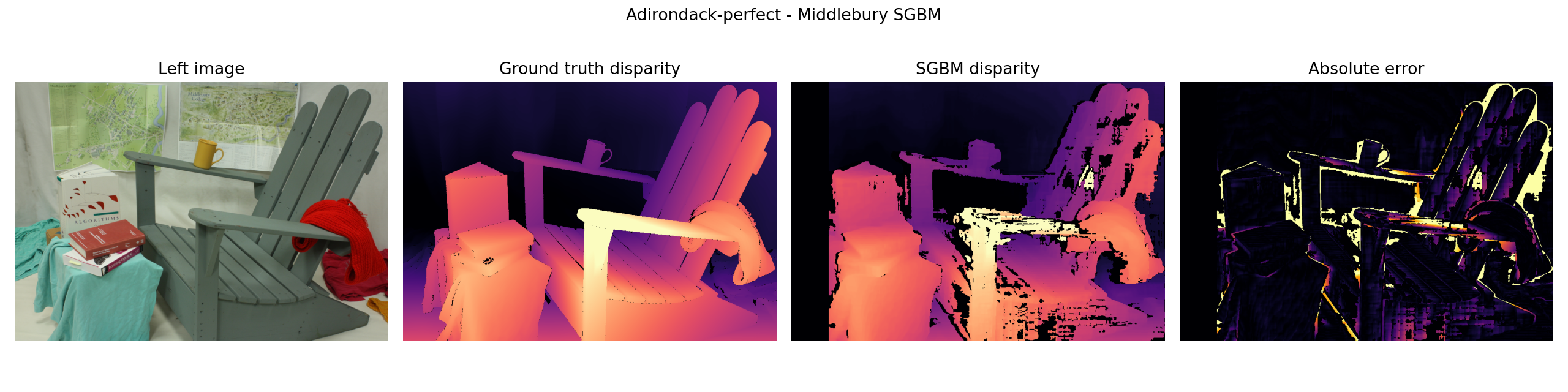}
    \includegraphics[width=\linewidth]{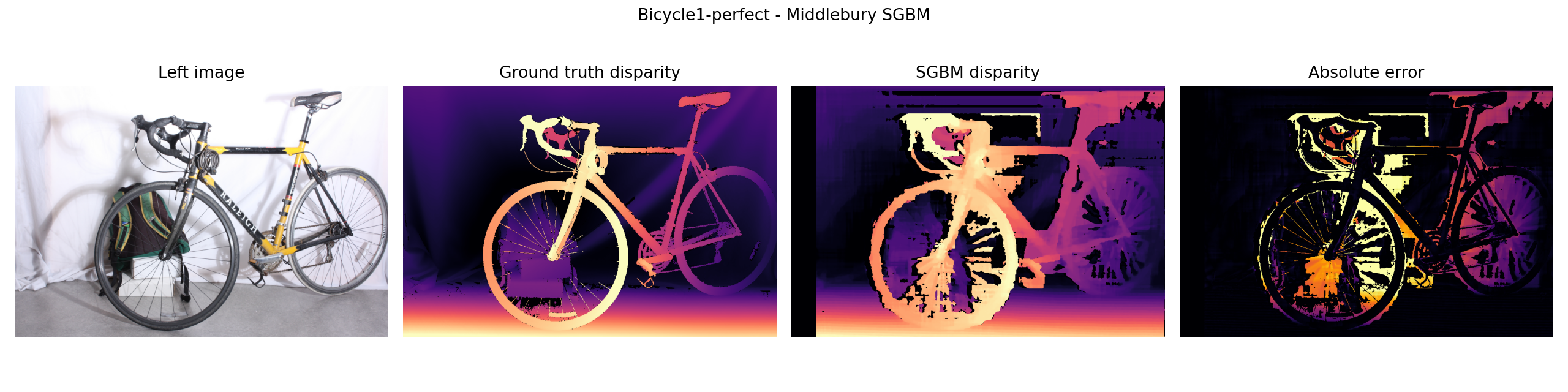}
    \includegraphics[width=\linewidth]{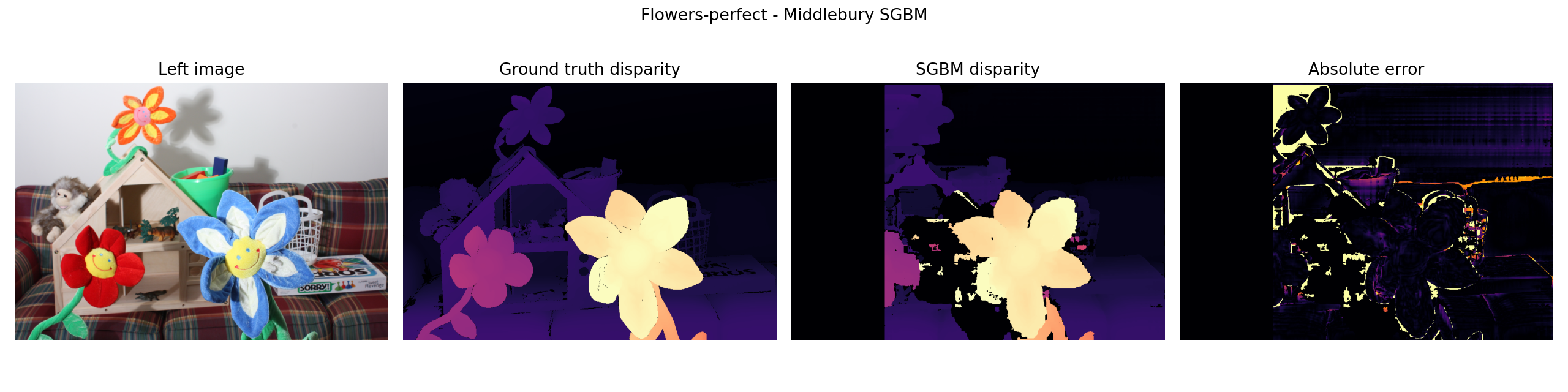}
    \includegraphics[width=\linewidth]{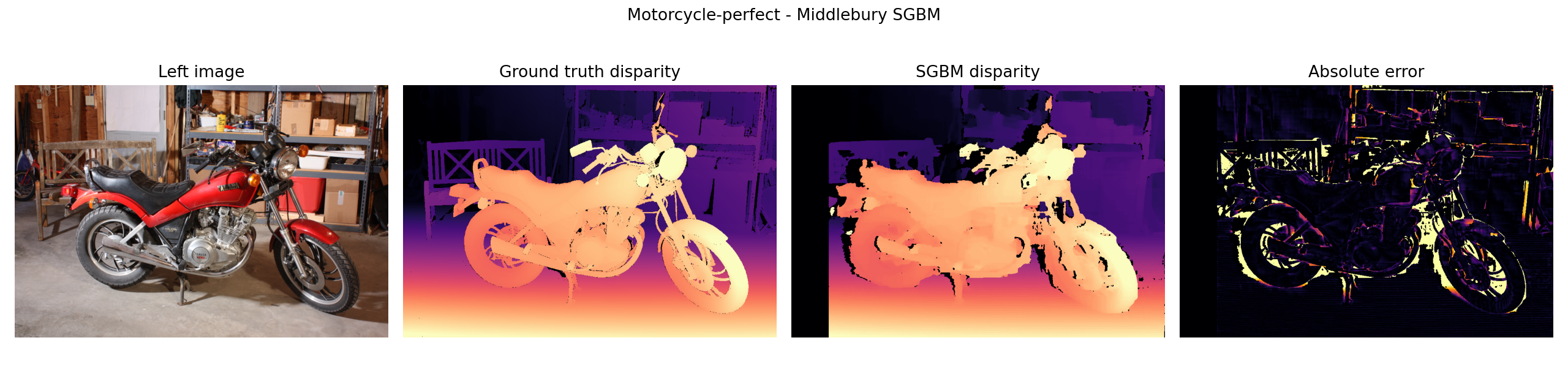}
    \includegraphics[width=\linewidth]{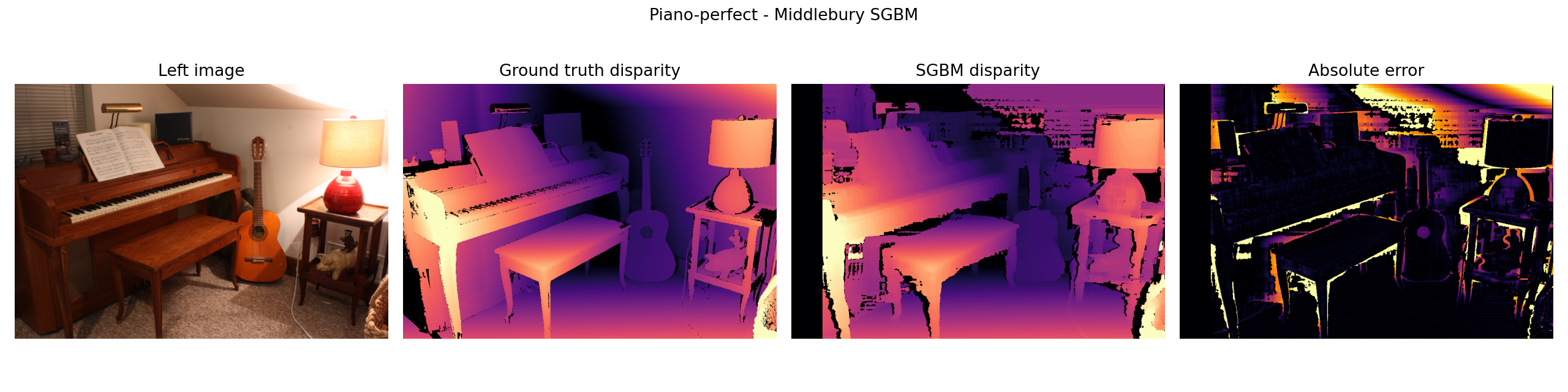}
    \caption{Ground-truth disparity vs. SGBM disparity results for Middlebury dataset (5 random samples).}
    \label{fig:middlebury_sgbm}
\end{figure}

\begin{figure}[htbp]
    \centering
    \includegraphics[width=\linewidth]{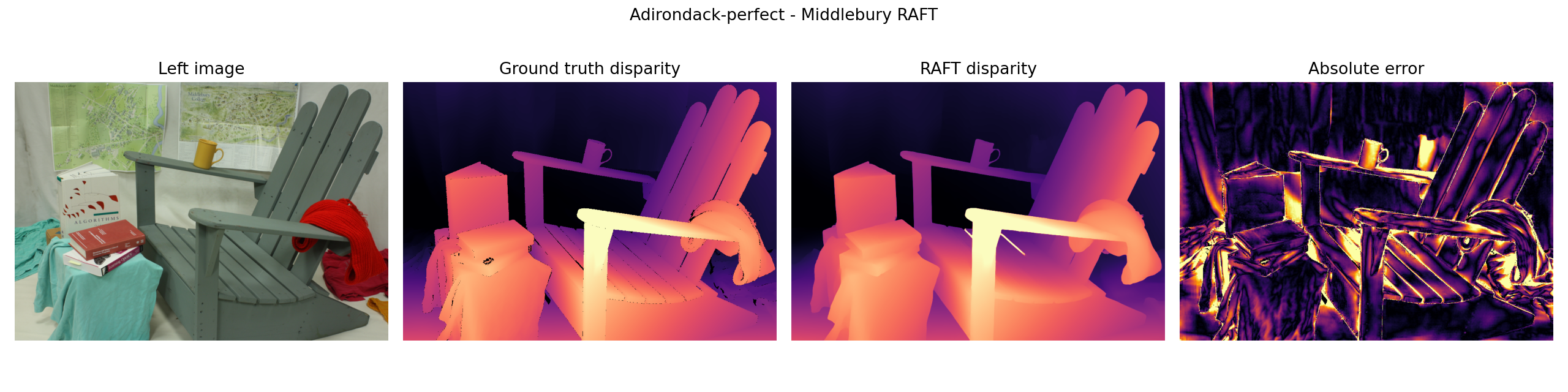}
    \includegraphics[width=\linewidth]{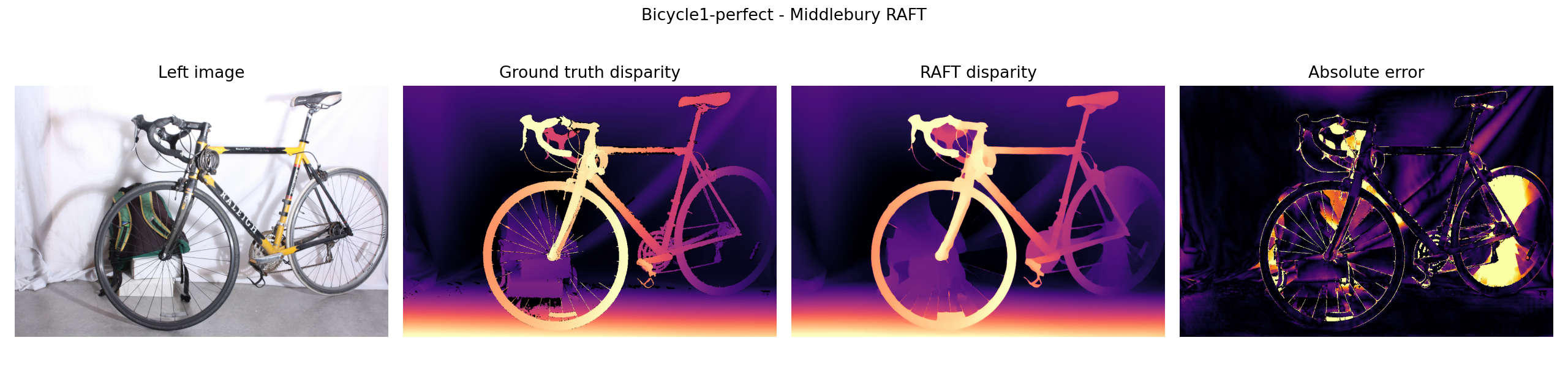}
    \includegraphics[width=\linewidth]{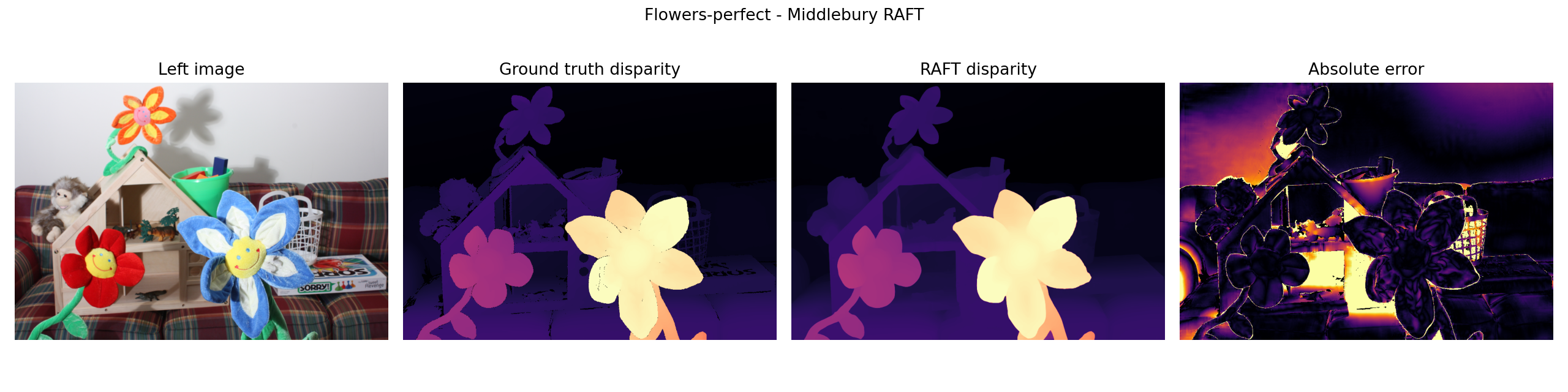}
    \includegraphics[width=\linewidth]{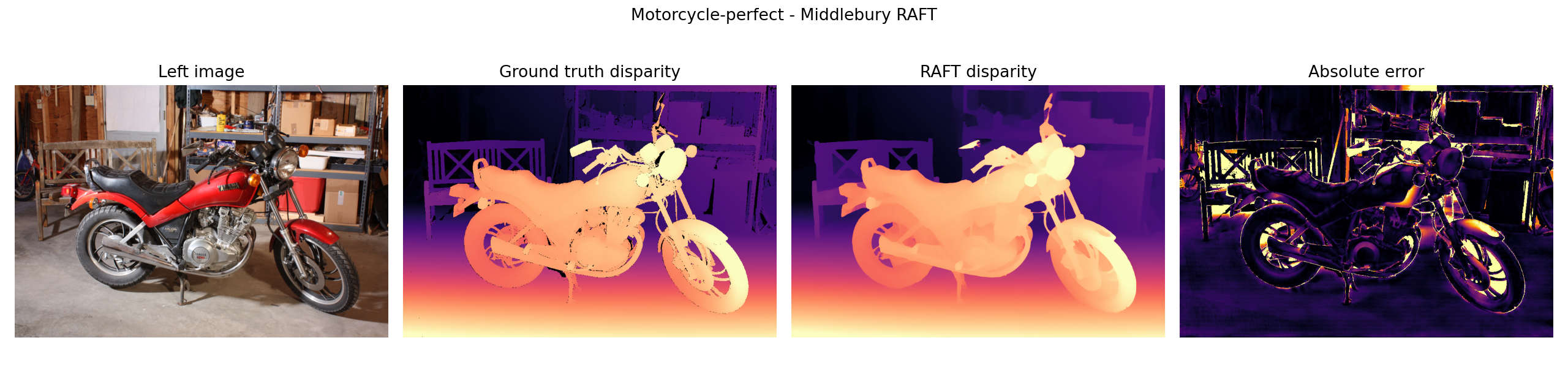}
    \includegraphics[width=\linewidth]{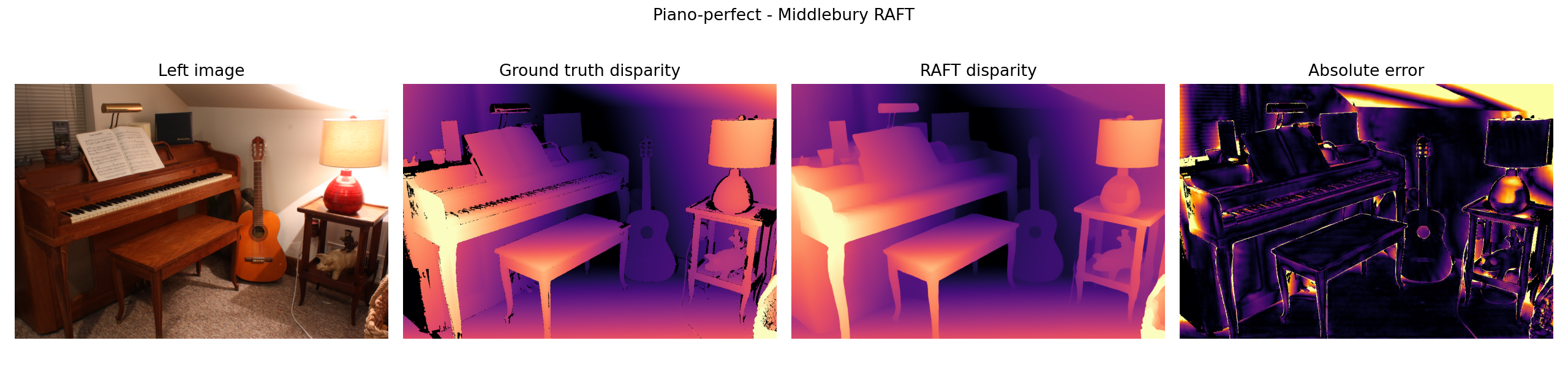}
    \caption{Ground-truth disparity vs. RAFT-Stereo disparity results for Middlebury dataset (5 random samples).}
    \label{fig:middlebury_raft}
\end{figure}

We first compared SGBM and RAFT-Stereo on Middlebury and Curiosity imagery to test whether performance on a standard stereo benchmark transferred to low-texture planetary scenes (Figures \ref{fig:middlebury_sgbm} and \ref{fig:middlebury_raft}). On Middlebury, RAFT-Stereo outperformed SGBM across all quantitative metrics (Table~\ref{tab:middlebury}), producing valid predictions for all evaluated pixels, reducing disparity MAE from 3.22\,px to 0.73\,px, and lowering the bad-4\,px rate from 12.6\% to 3.1\%. SGBM recovered valid disparities for 76.3\% of ground-truth pixels, with larger errors near occlusions, discontinuities, and weakly textured regions.

On Curiosity imagery, the relative behavior of the two methods changed (Table~\ref{tab:curiosity_proxy}). RAFT-Stereo had valid disparities over 99.8\% of pixels and valid depths over 96.9\% after calibrated depth conversion, but slightly higher photometric reprojection error than SGBM (0.166 vs. 0.153). SGBM retained significantly fewer pixels.

\begin{figure}[htbp]
    \centering
    \includegraphics[width=\linewidth]{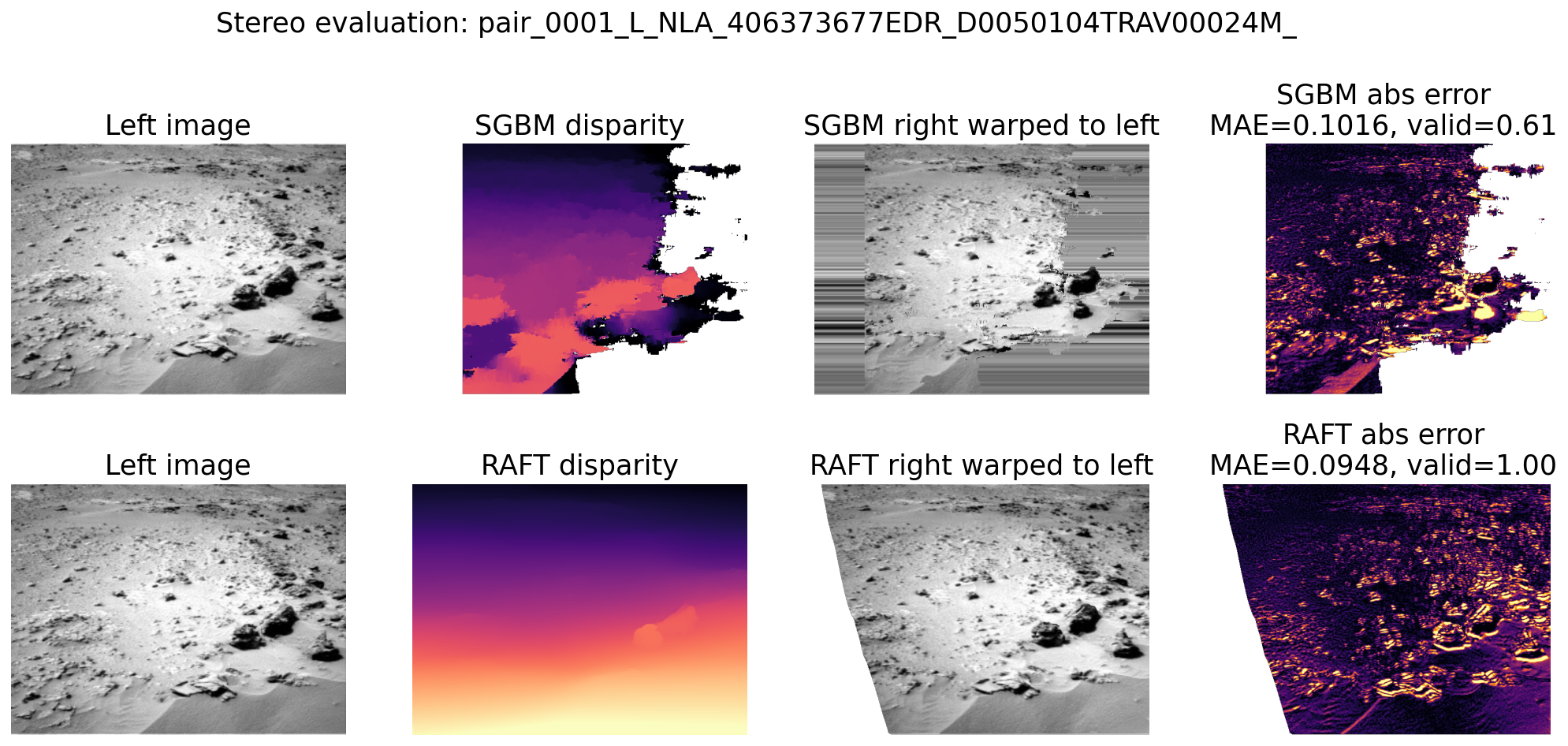}
    \includegraphics[width=\linewidth]{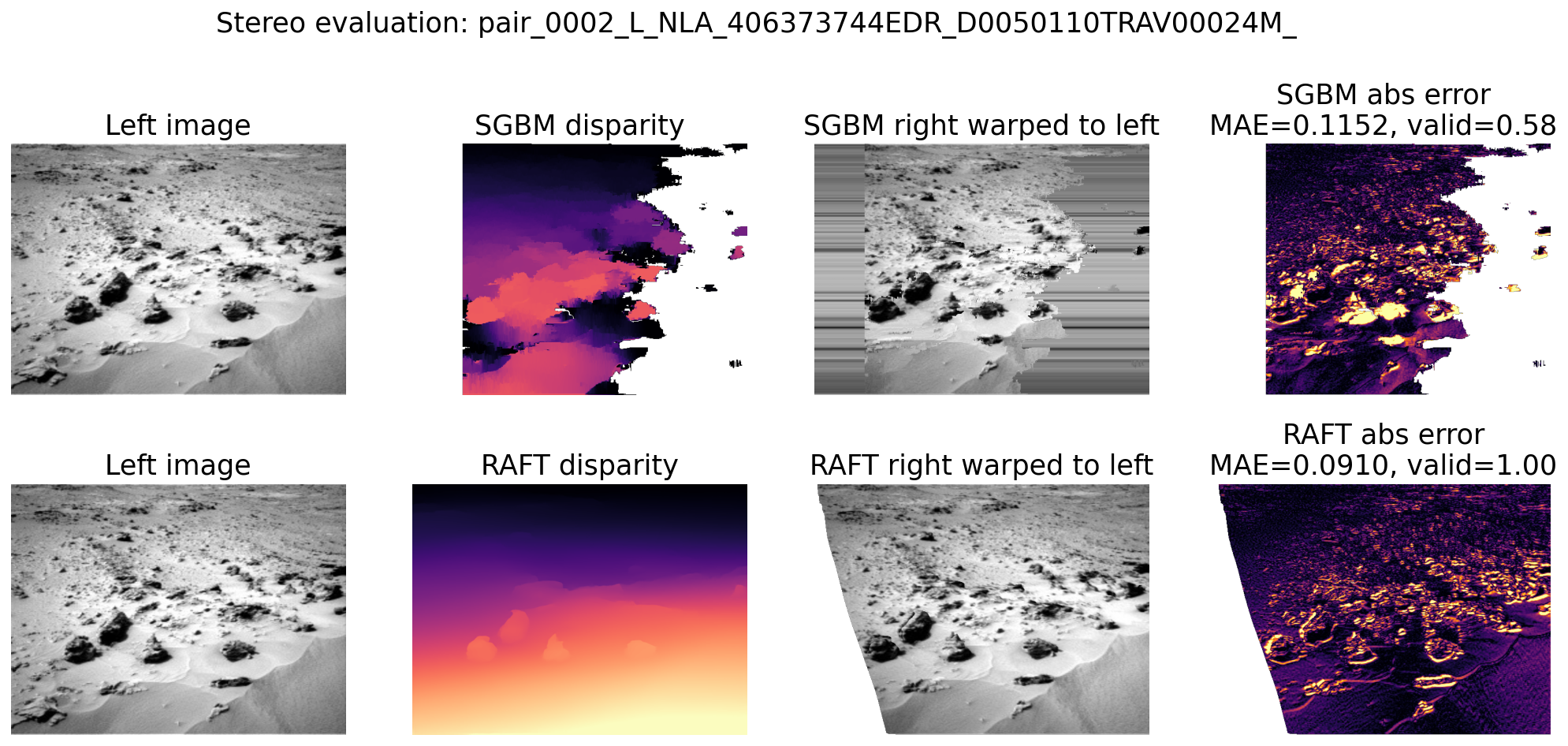}
    \includegraphics[width=\linewidth]{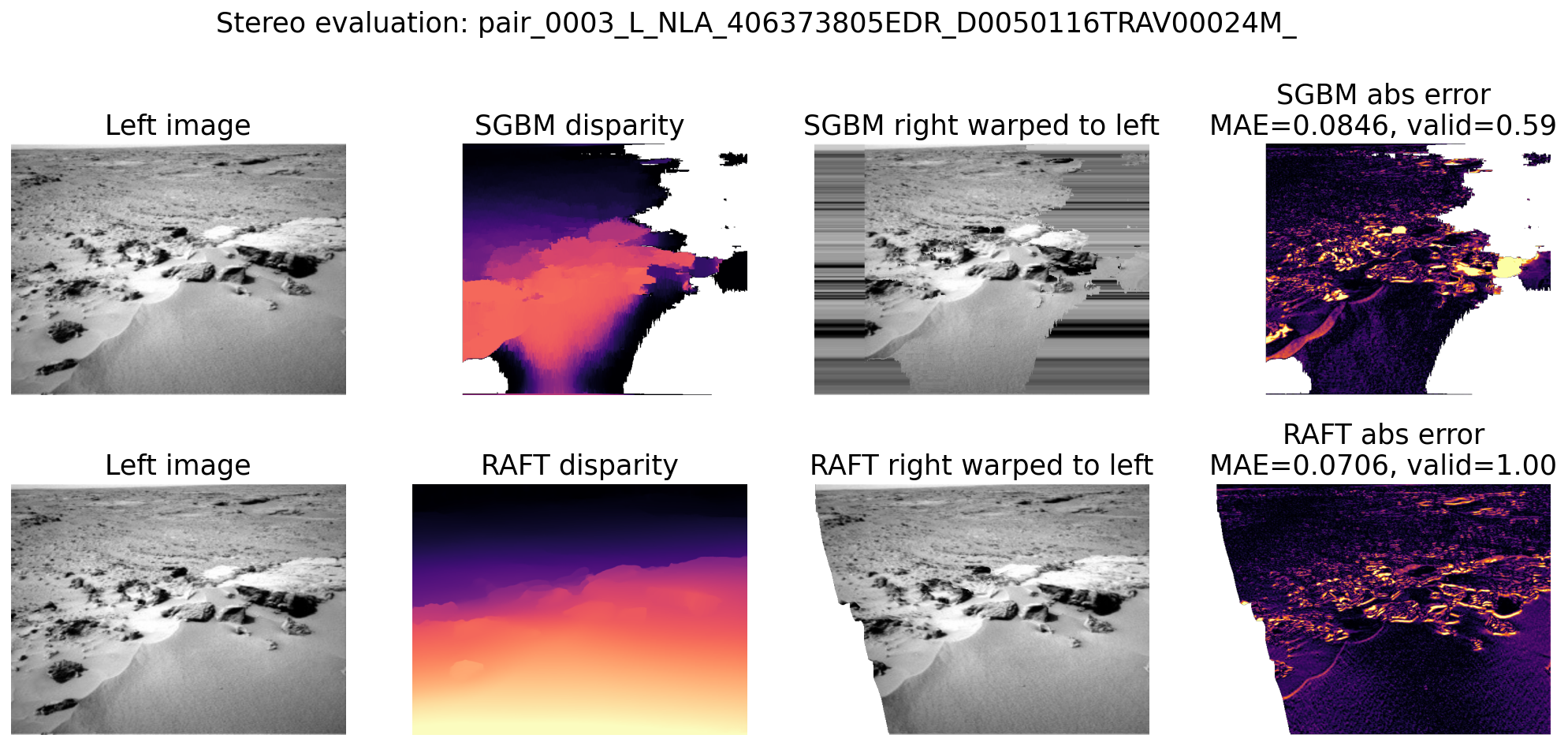}
    \includegraphics[width=\linewidth]{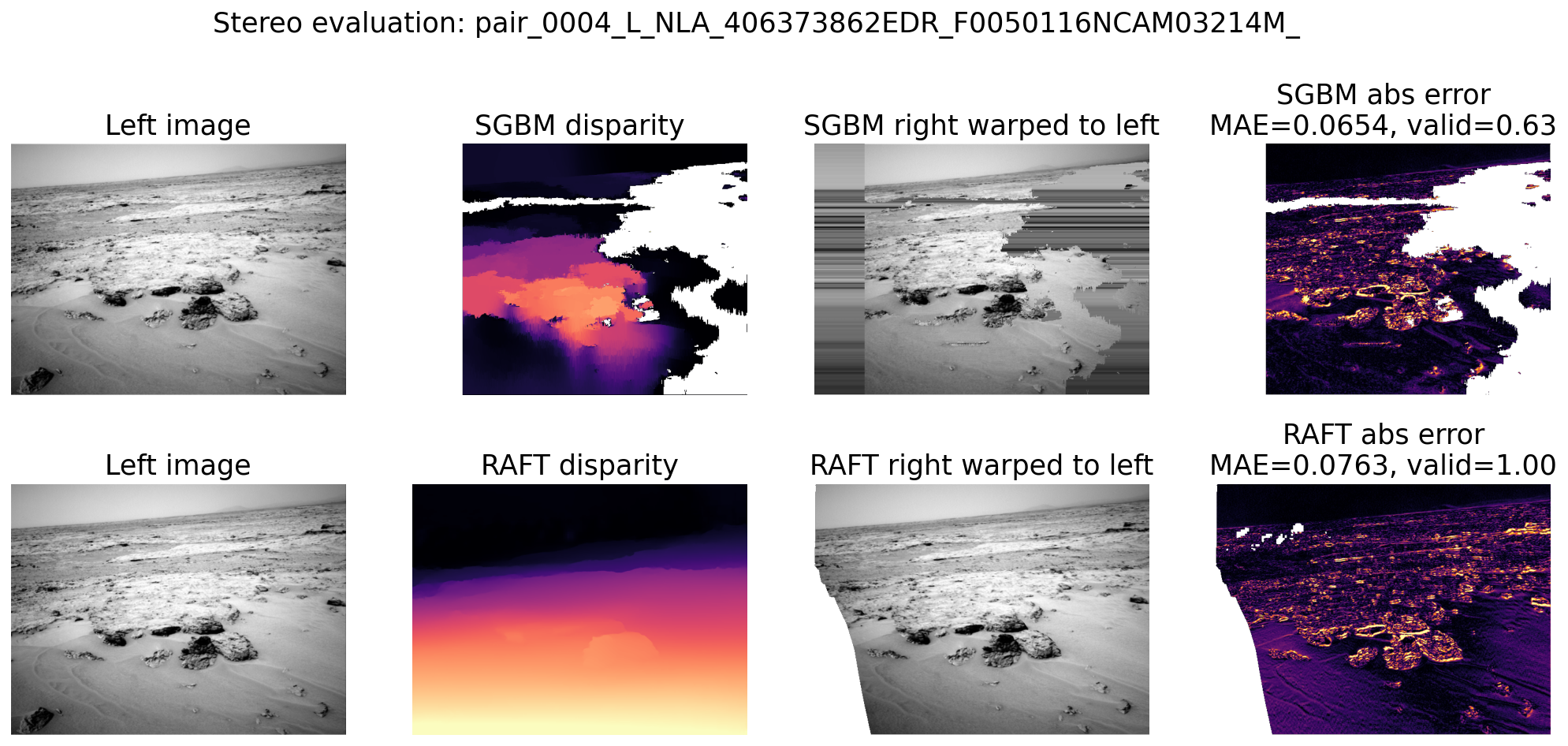}
    \includegraphics[width=\linewidth]{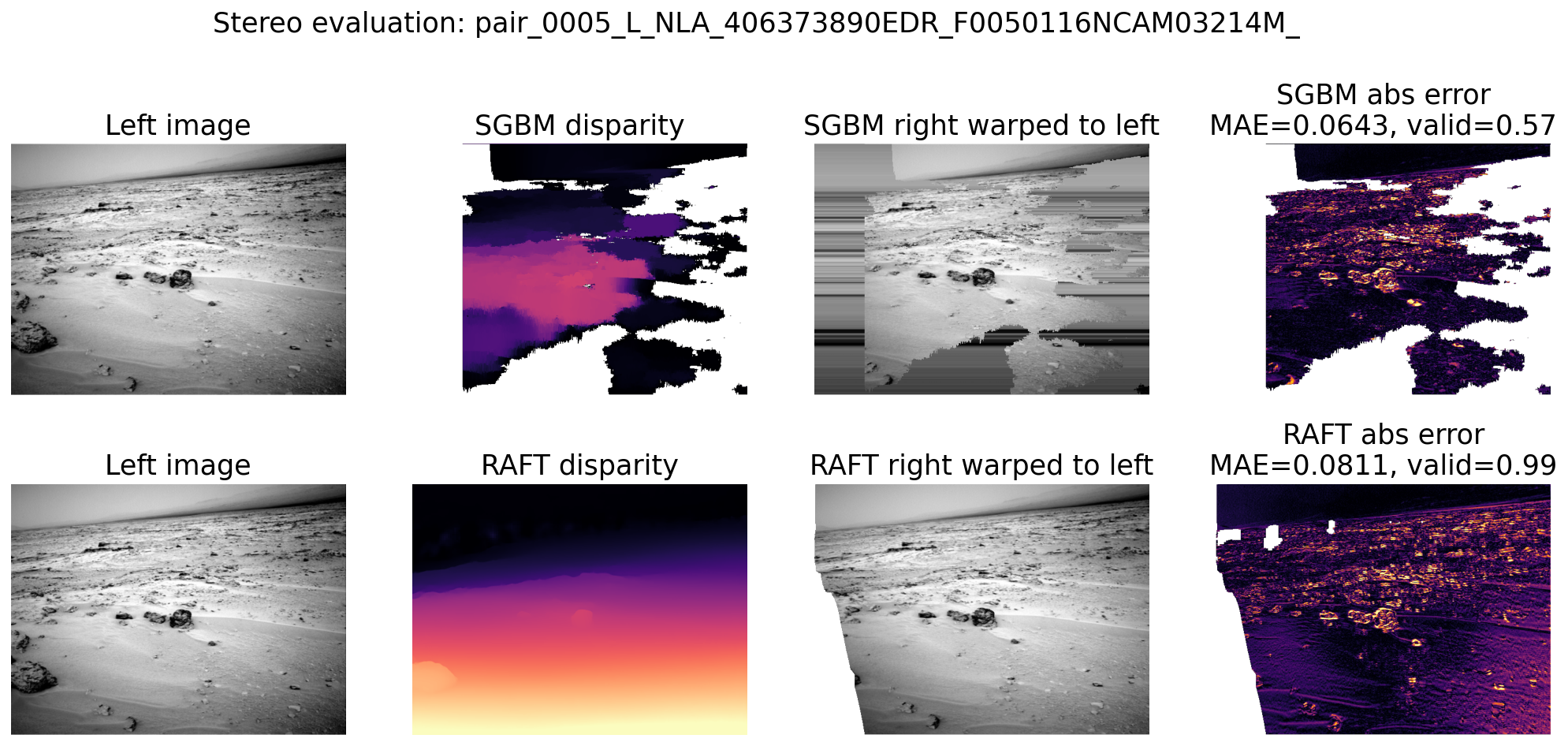}
    \caption{SGBM vs. RAFT-Stereo disparity for Curiosity rover images (first 5 samples).}
    \label{fig:curiosity_sgbm_raft}
\end{figure}

However, density and reprojection consistency did not fully capture terrain fidelity. SGBM achieved stronger edge alignment, with an image-gradient/disparity-gradient correlation of 0.174 compared to 0.130 for RAFT. Figure~\ref{fig:curiosity_sgbm_raft} shows the same pattern qualitatively.

RAFT also produced larger median disparities and shallower median depths, with a median depth of 0.862\,m compared to 1.759\,m for SGBM (Table~\ref{tab:curiosity_stats}). This likely reflects the different disparity supports. The shallower RAFT depths can therefore be interpreted as evidence of domain shift rather than direct evidence that the terrain was closer.

\begin{table}[htbp]
    \centering
    \small
    \setlength{\tabcolsep}{4pt}
    \caption{Middlebury stereo reconstruction accuracy. Lower is better for MAE, RMSE, and bad-pixel rates. Bad-1 and Bad-4 denote the fraction of evaluated pixels with absolute disparity error greater than 1\,px and 4\,px, respectively.}
    \label{tab:middlebury}
    \begin{tabular}{lccccc}
        \toprule
        Method & Valid & MAE & RMSE & Bad-1 & Bad-4 \\
        \midrule
        SGBM & 0.763 & 3.224 & 9.892 & 0.182 & 0.126 \\
        RAFT-Stereo & 1.000 & 0.729 & 3.049 & 0.111 & 0.031 \\
        \bottomrule
    \end{tabular}
\end{table}

\begin{table}[htbp]
    \centering
    \small
    \setlength{\tabcolsep}{4pt}
    \caption{Curiosity stereo metrics after calibrated depth conversion. Valid disp. is the fraction of pixels with finite positive disparity. Valid depth is the fraction with finite depth in the accepted range. Reproj. is the fraction of pixels that remain valid after warping the right image into the left view. Photo. is the photometric MAE after reprojection. Edge is the correlation between image gradients and disparity gradients.}
    \label{tab:curiosity_proxy}
    \begin{tabular}{lccccc}
        \toprule
        Method & Disp. & Depth & Reproj. & Photo. & Edge \\
        \midrule
        SGBM & 0.560 & 0.524 & 0.519 & 0.153 & 0.174 \\
        RAFT-Stereo & 0.998 & 0.969 & 0.748 & 0.166 & 0.130 \\
        \bottomrule
    \end{tabular}
\end{table}

\begin{table}[htbp]
    \centering
    \small
    \setlength{\tabcolsep}{4pt}
    \caption{Curiosity disparity and depth statistics after calibrated depth conversion. Values report the average per-scene median disparity, median depth, and median absolute disparity gradient.}
    \label{tab:curiosity_stats}
    \begin{tabular}{lccc}
        \toprule
        Method & Median disp. & Median depth & Disp. grad. \\
        \midrule
        SGBM & 44.0 & 1.759 & 0.098 \\
        RAFT-Stereo & 81.1 & 0.862 & 0.147 \\
        \bottomrule
    \end{tabular}
\end{table}

Taken together, the stereo results revealed a domain-shift gap. RAFT-Stereo performed better on Middlebury, but this benchmark advantage did not translate to more reliable Martian terrain reconstruction. On Curiosity images, standard metrics favored RAFT in density and reprojection-valid coverage, whereas photometric error, edge alignment, and depth statistics revealed additional failure modes.

\subsection{Occlusion estimation}

\begin{figure}[htbp]
    \centering
    \includegraphics[width=\linewidth]{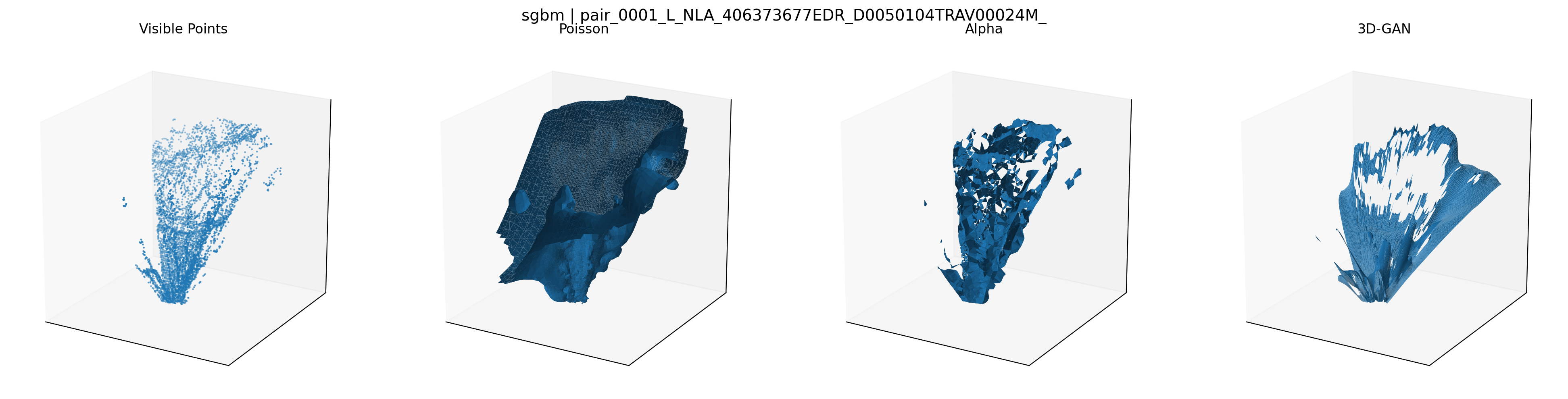}
    \includegraphics[width=\linewidth]{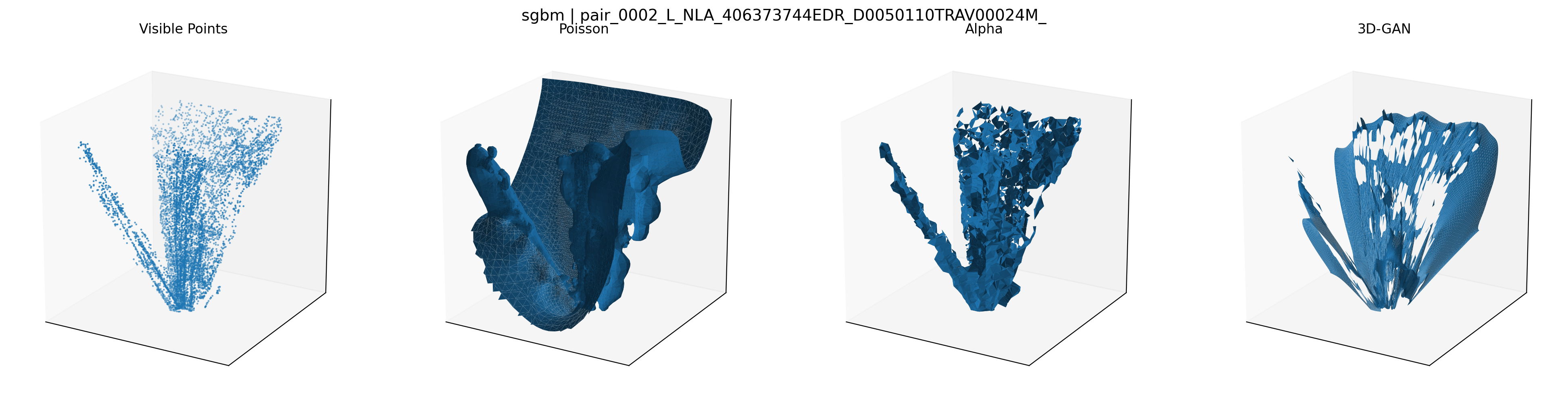}
    \includegraphics[width=\linewidth]{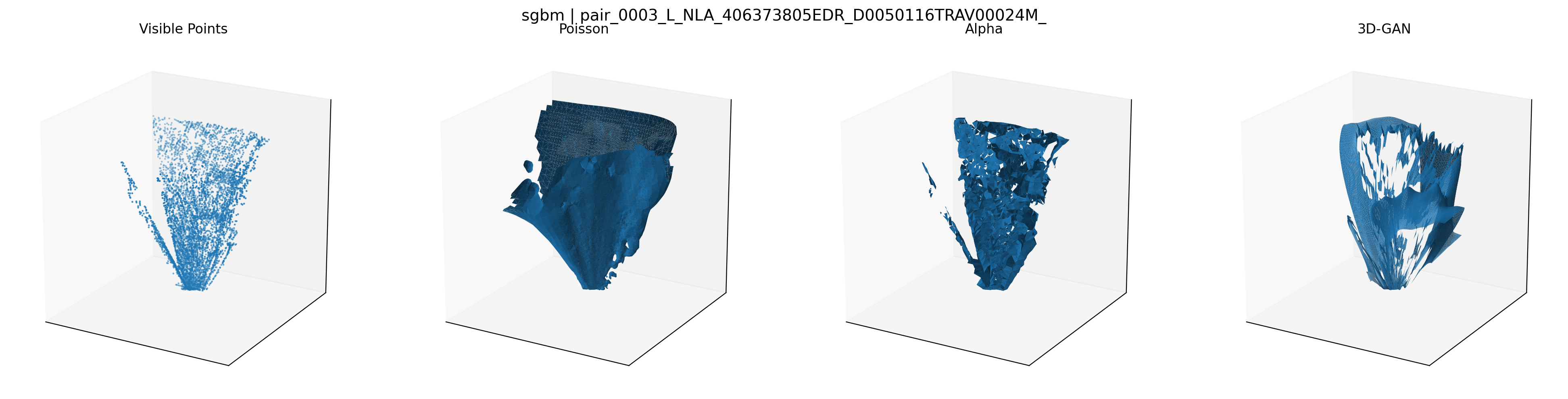}
    \includegraphics[width=\linewidth]{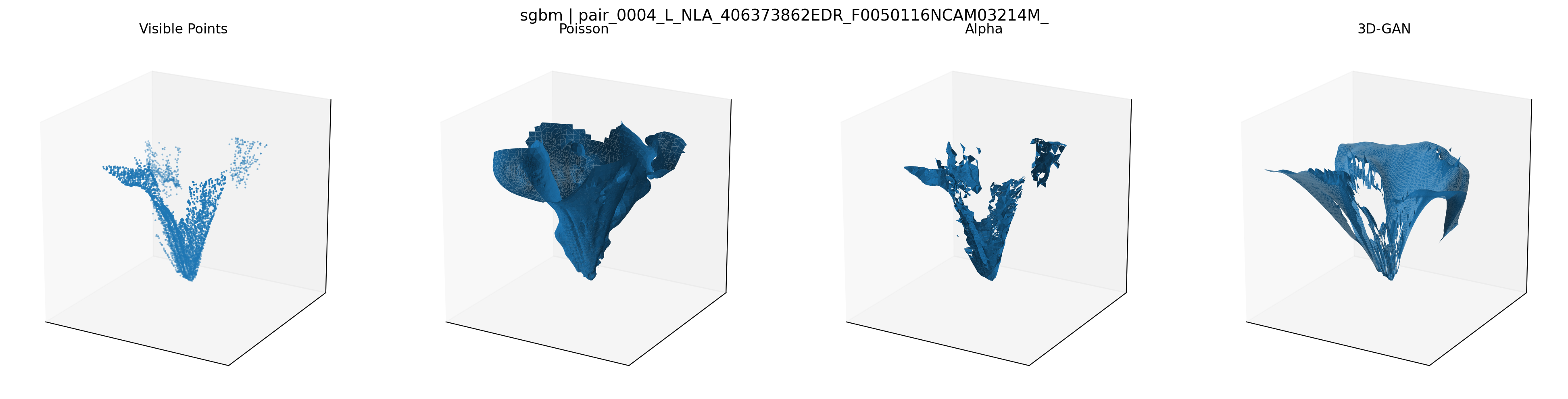}
    \includegraphics[width=\linewidth]{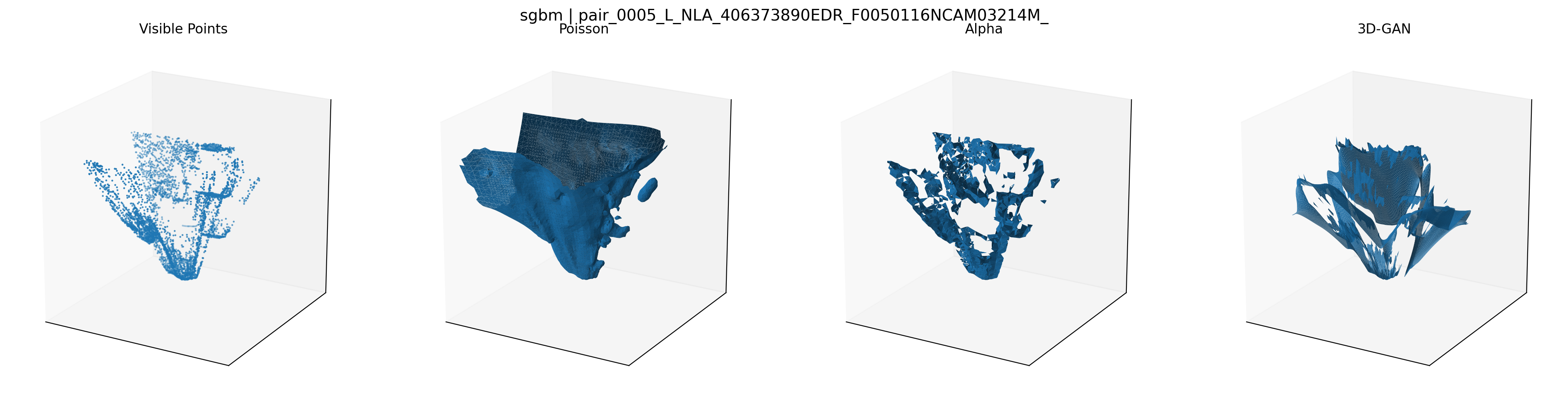}
    \caption{SGBM-derived geometry completion examples showing visible points, Poisson, alpha-shape, and deterministic diffusion-fill outputs.}
    \label{fig:3d_sgbm}
\end{figure}

\begin{figure}[htbp]
    \centering
    \includegraphics[width=\linewidth]{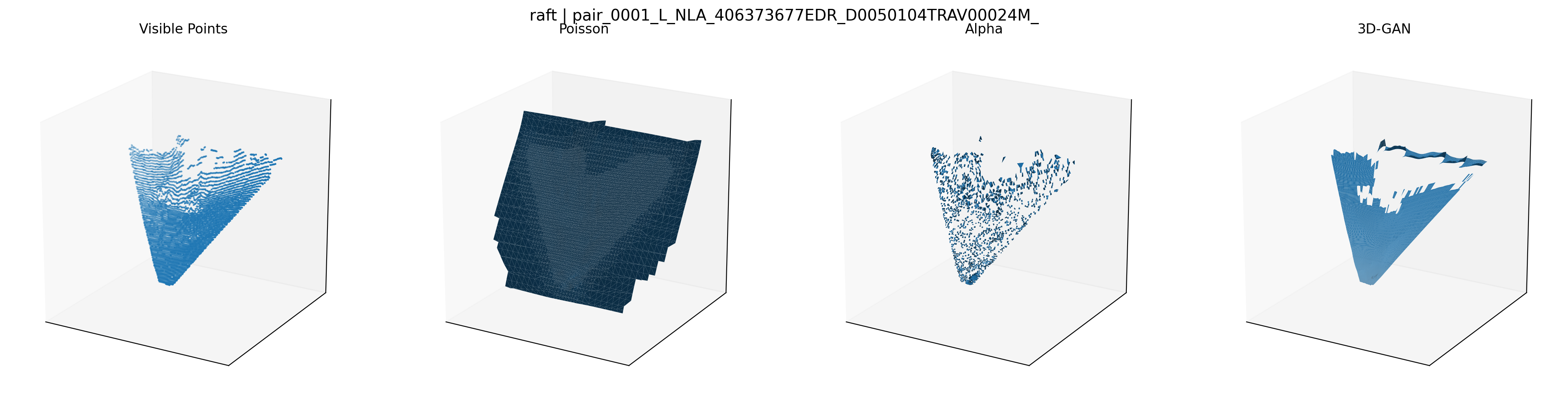}
    \includegraphics[width=\linewidth]{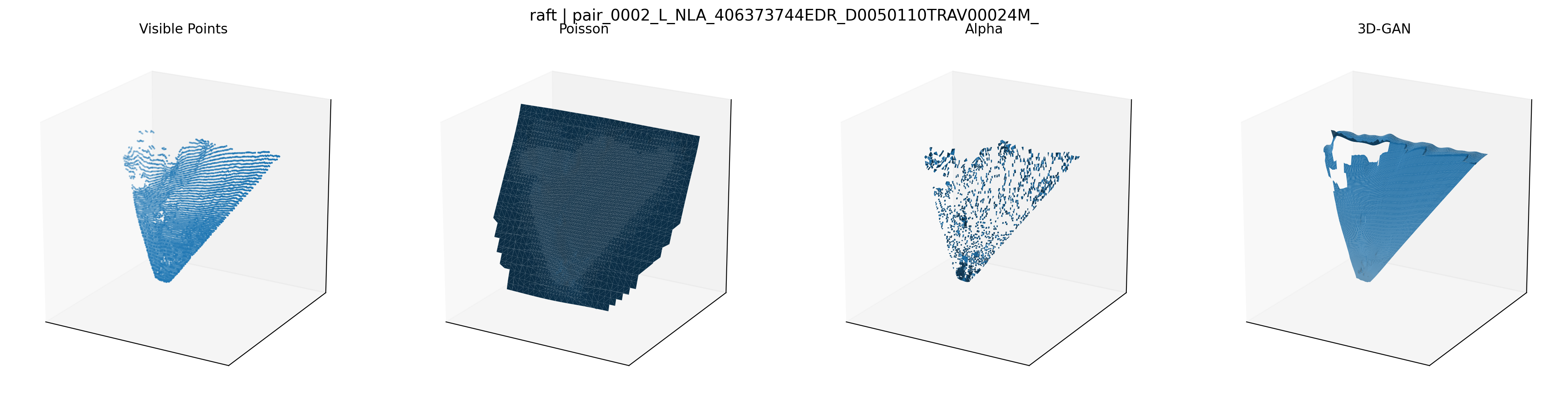}
    \includegraphics[width=\linewidth]{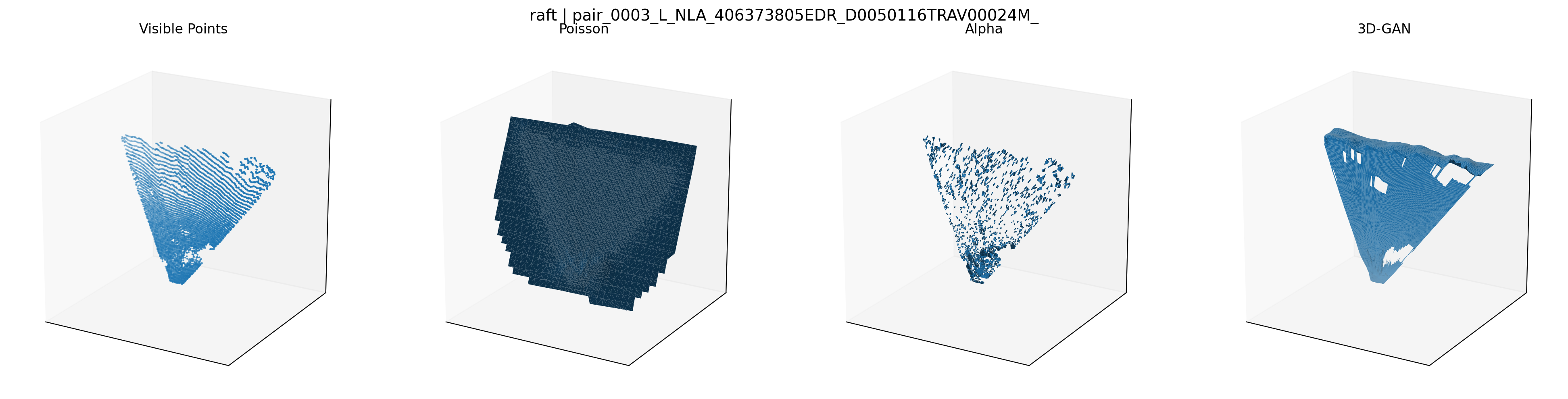}
    \includegraphics[width=\linewidth]{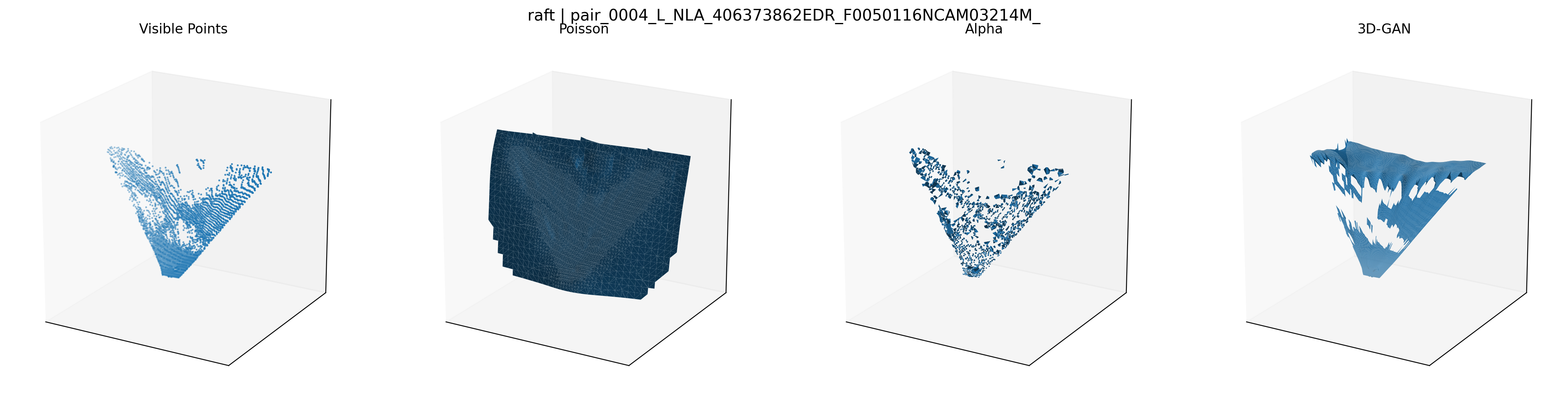}
    \includegraphics[width=\linewidth]{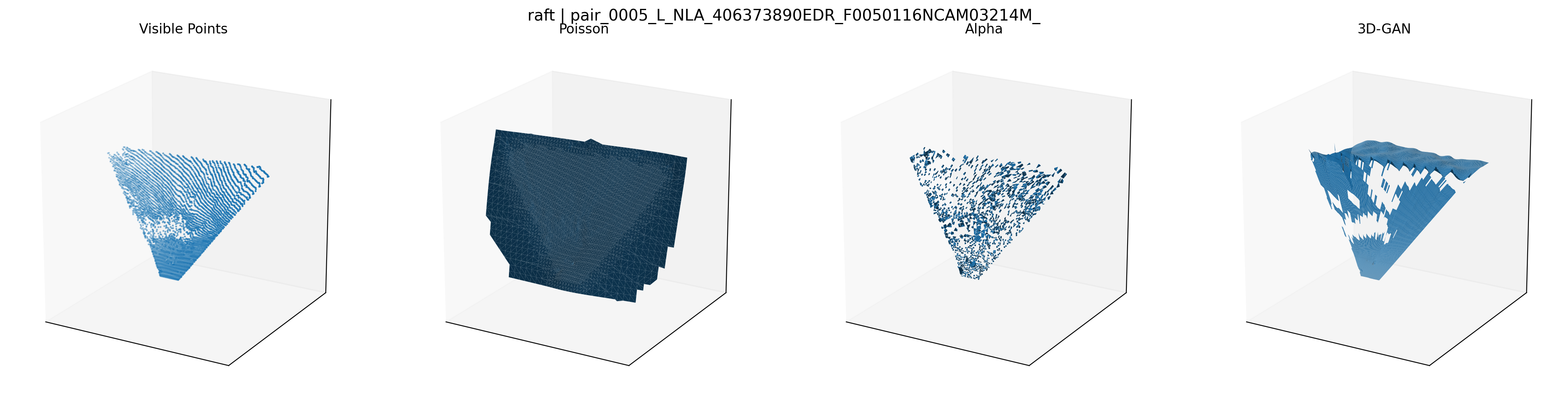}
    \caption{RAFT-derived geometry completion examples showing visible points, Poisson, alpha-shape, and deterministic diffusion-fill outputs.}
    \label{fig:3d_raft}
\end{figure}

After identifying these stereo-domain effects, we evaluated how the resulting depth errors propagated into occluded-surface estimation. Completion demonstrated a tradeoff between fidelity to the observed point cloud and topological connectivity (Figures~\ref{fig:3d_sgbm} and \ref{fig:3d_raft}).

Alpha-shape reconstruction preserved the measured surface most closely. All completion methods produced valid outputs for the 100 SGBM-derived and 100 RAFT-derived reconstructions. Alpha shapes achieved the lowest normalized Chamfer errors and high visible-surface coverage, indicating strong agreement with the observed point cloud (Table~\ref{tab:completion_fidelity}).

This local fidelity came at the cost of global mesh structure. SGBM-alpha meshes had a largest-component fraction of 0.799, but RAFT-alpha meshes were highly fragmented, averaging 894 connected components and a largest-component fraction of only 0.100 (Table~\ref{tab:completion_topology}).

Poisson reconstruction followed the opposite pattern. It produced more connected surfaces, especially for RAFT inputs, where the largest-component fraction reached 0.996 with only 1.61 connected components on average. The cost of this connectivity was reduced agreement with the observed surface. The deterministic diffusion-fill baseline produced an intermediate result for RAFT, but it was less reliable for SGBM inputs.

\begin{table}[htbp]
    \centering
    \small
    \setlength{\tabcolsep}{3.5pt}
    \caption{Geometry completion fidelity on Curiosity reconstructions after calibrated depth filtering. Cham. is normalized Chamfer distance to the visible point cloud. Cov. is the fraction of visible points within the distance threshold. Nov. is the fraction of mesh samples farther than the threshold from the visible surface. Lower Cham. and Nov. indicate closer agreement with the observed surface.}
    \label{tab:completion_fidelity}
    \begin{tabular}{llccc}
        \toprule
        Stereo & Comp. & Cham. & Cov. & Nov. \\
        \midrule
        SGBM & Alpha & 0.0127 & 0.904 & 0.181 \\
        SGBM & Poisson & 0.0553 & 0.698 & 0.741 \\
        SGBM & Diff.-fill & 0.0592 & 0.591 & 0.697 \\
        RAFT & Alpha & 0.0089 & 0.874 & 0.015 \\
        RAFT & Poisson & 0.0650 & 0.901 & 0.605 \\
        RAFT & Diff.-fill & 0.0182 & 0.914 & 0.149 \\
        \bottomrule
    \end{tabular}
\end{table}

\begin{table}[htbp]
    \centering
    \small
    \setlength{\tabcolsep}{3.5pt}
    \caption{Geometry completion topology on Curiosity reconstructions after calibrated depth filtering. Vert. and Tri. are the average numbers of mesh vertices and triangles. CC is the average number of connected components. LCC is the fraction of triangles in the largest connected component. Lower CC and higher LCC indicate more connected surfaces.}
    \label{tab:completion_topology}
    \begin{tabular}{llcccc}
        \toprule
        Stereo & Comp. & Vert. & Tri. & CC & LCC \\
        \midrule
        SGBM & Alpha & 3.3k & 6.7k & 69.5 & 0.799 \\
        SGBM & Poisson & 14.1k & 28.1k & 21.4 & 0.973 \\
        SGBM & Diff.-fill & 11.8k & 22.1k & 35.8 & 0.946 \\
        RAFT & Alpha & 4.1k & 5.6k & 893.8 & 0.100 \\
        RAFT & Poisson & 9.8k & 19.4k & 1.6 & 0.996 \\
        RAFT & Diff.-fill & 12.2k & 23.7k & 4.0 & 0.988 \\
        \bottomrule
    \end{tabular}
\end{table}

These results show that no completion method produced both accurate and connected reconstructions on its own. Alpha shapes preserved accurate but fragmented geometry, Poisson reconstruction produced smoother, more connected meshes by adding unsupported geometry, and the deterministic diffusion-fill baseline depended strongly on the stereo input. Figures~\ref{fig:3d_sgbm} and \ref{fig:3d_raft} show this comparison visually.

\subsection{Completion ablation}

The synthetic-ground-truth ablation showed that the fidelity-connectivity tradeoff persisted even when stereo error was removed (Table~\ref{tab:completion_ablation}). The three completion methods produced valid reconstructions for all 12 synthetic cases. Across four synthetic terrain surfaces with known geometry and three structured occlusion patterns, the deterministic diffusion-fill baseline achieved the best average geometric agreement with the known surface, with the lowest normalized Chamfer distance (0.0117) and the highest ground-truth coverage (0.840). Poisson reconstruction remained fully connected on every synthetic case and achieved intermediate fidelity (Chamfer 0.0370, coverage 0.755), but its higher novelty score (0.324) suggests oversmoothing or extrapolation beyond the ground-truth surface. Alpha shapes introduced essentially no novel geometry, but they fragmented severely, averaging 780 connected components and only 0.304 coverage of the ground-truth surface.

\begin{table}[htbp]
    \centering
    \small
    \setlength{\tabcolsep}{3.5pt}
    \caption{Synthetic-ground-truth completion ablation. Cham. is normalized Chamfer distance to sampled ground-truth surface points. Cov. is the fraction of ground-truth surface points recovered within the evaluation threshold. Nov. is the fraction of mesh samples farther than the threshold from the ground-truth surface. Lower Cham. and Nov. indicate better geometric fidelity.}
    \label{tab:completion_ablation}
    \begin{tabular}{lccccc}
        \toprule
        Comp. & Cham. & Cov. & Nov. & CC & LCC \\
        \midrule
        Alpha & 0.0579 & 0.304 & 0.000 & 779.6 & 0.013 \\
        Poisson & 0.0370 & 0.755 & 0.324 & 1.0 & 1.000 \\
        Diff.-fill & 0.0117 & 0.840 & 0.110 & 1.0 & 1.000 \\
        \bottomrule
    \end{tabular}
\end{table}

\subsection{Printable 3D mesh}

\begin{figure}[htbp!]
    \centering
    \includegraphics[width=\linewidth]{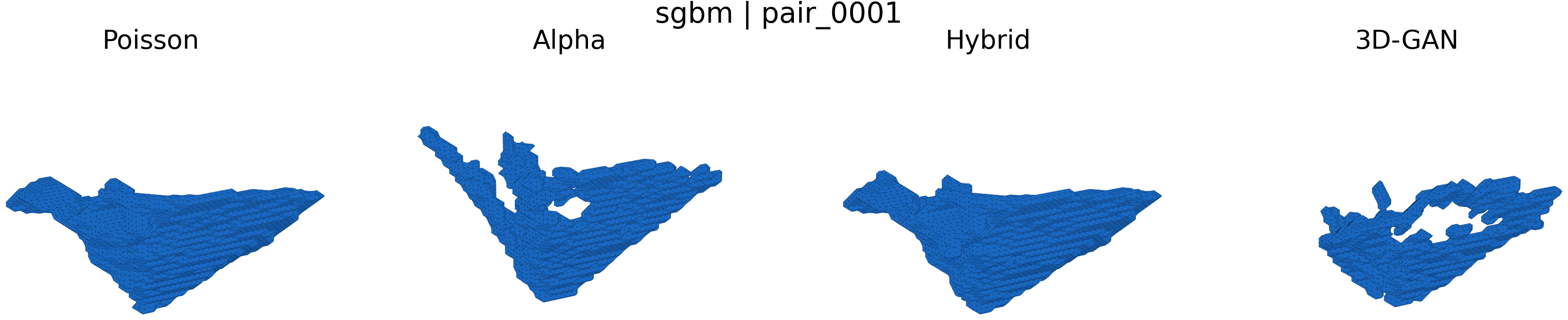}
    \includegraphics[width=\linewidth]{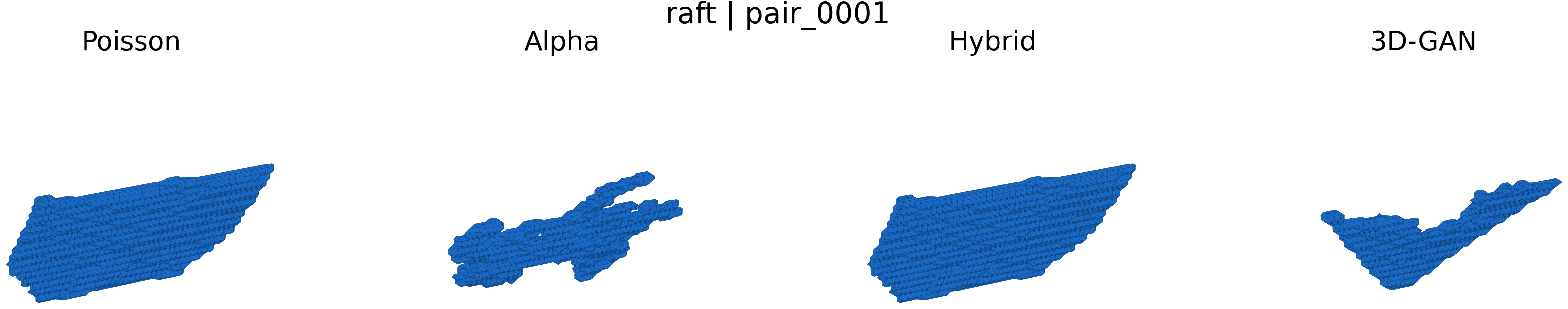}
    \includegraphics[width=\linewidth]{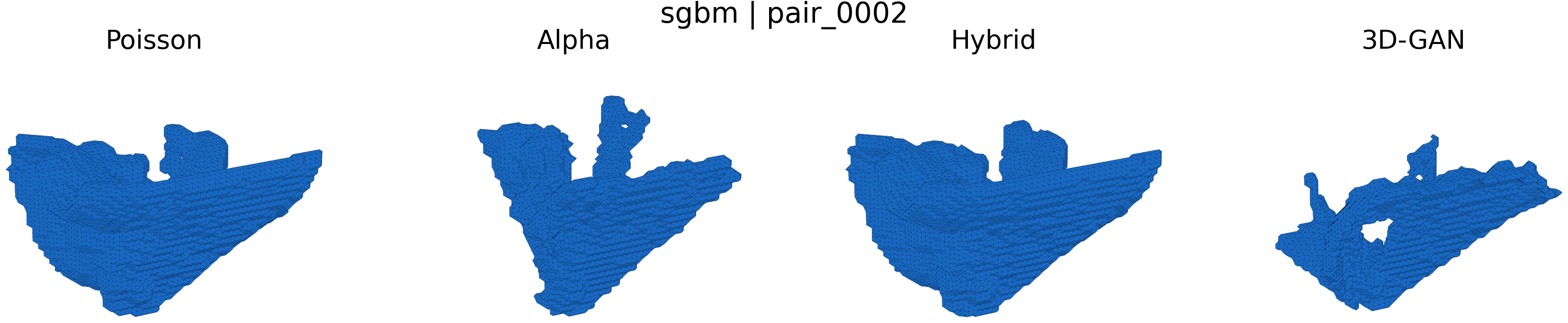}
    \includegraphics[width=\linewidth]{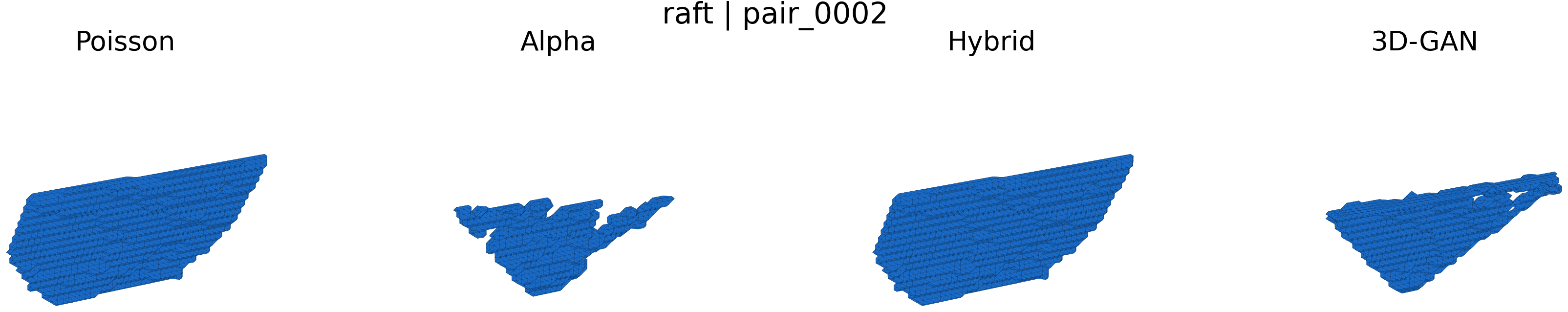}
    \includegraphics[width=\linewidth]{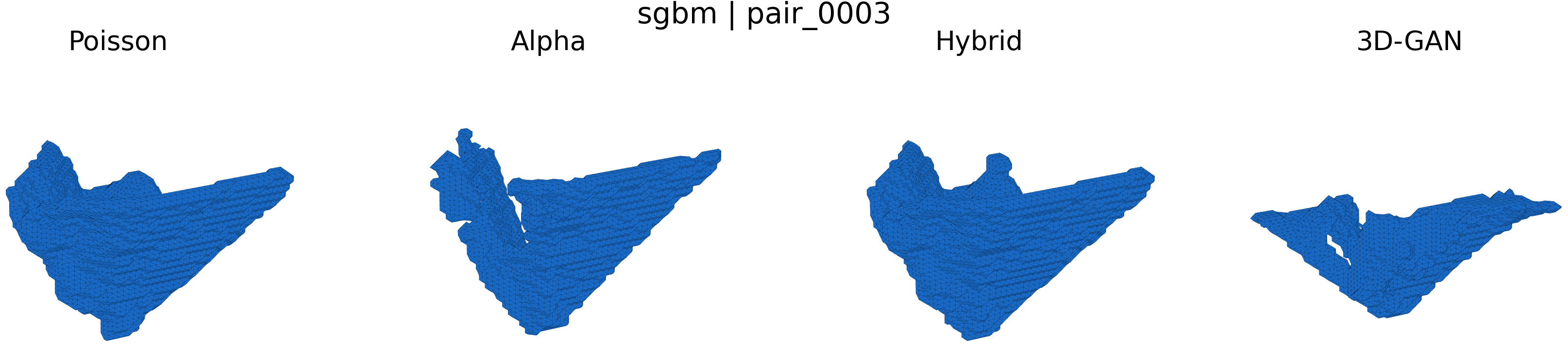}
    \includegraphics[width=\linewidth]{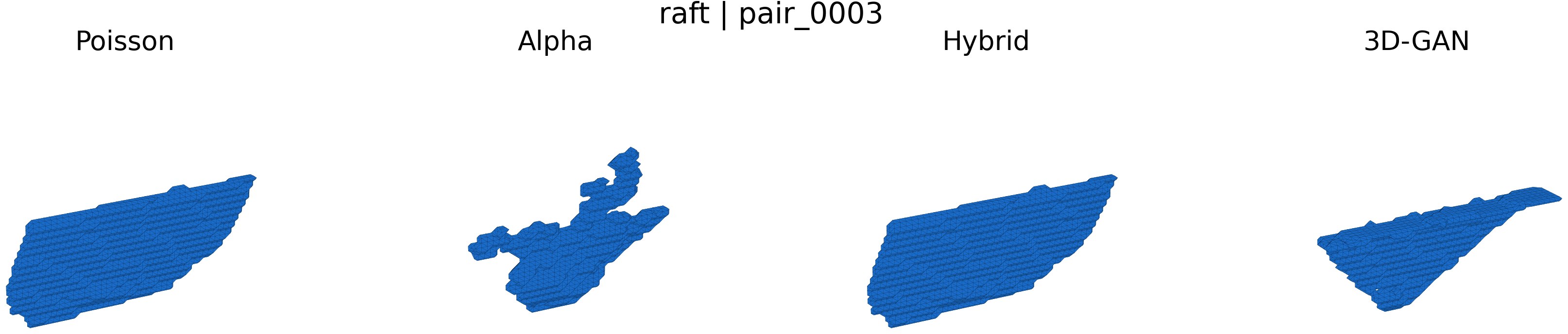}
    \includegraphics[width=\linewidth]{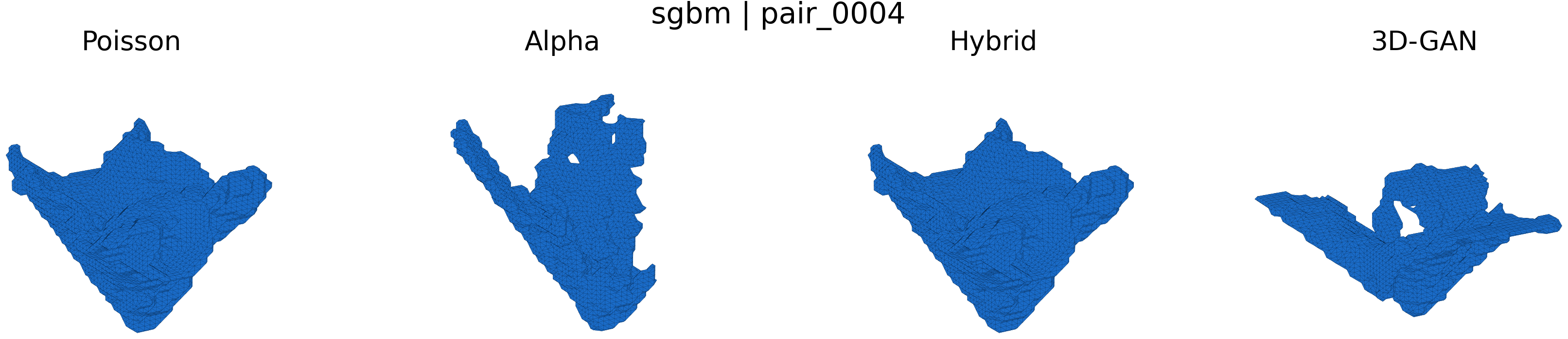}
    \includegraphics[width=\linewidth]{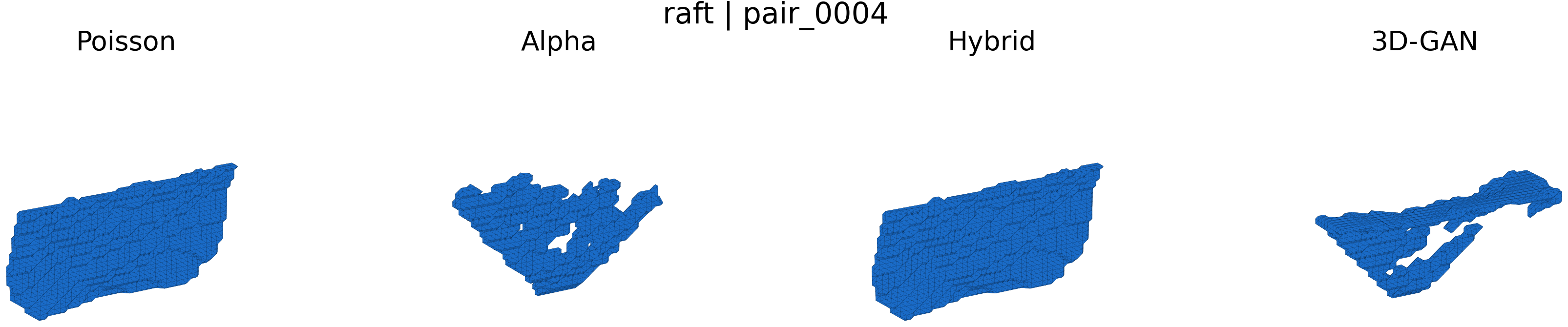}
    \includegraphics[width=\linewidth]{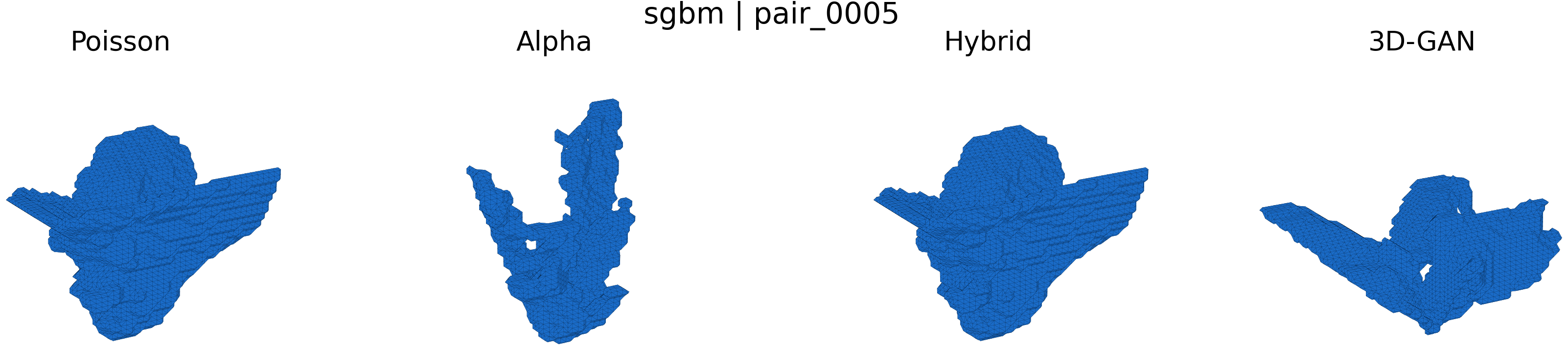}
    \includegraphics[width=\linewidth]{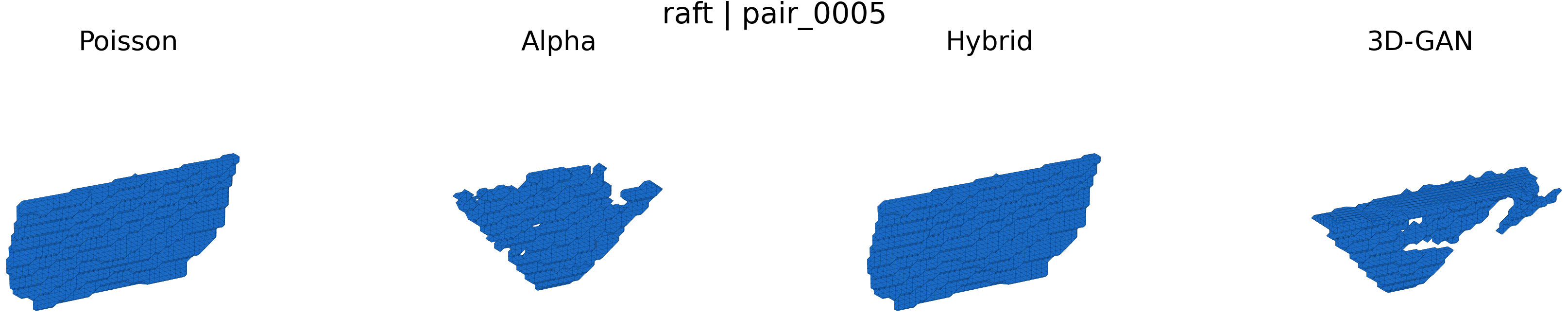}
    \caption{Printable meshes reconstructed from Curiosity stereo pairs after calibrated depth conversion. These examples visualize the final printable outputs. Each row compares Poisson, alpha-shape, hybrid, and deterministic diffusion-fill printable OBJ variants for either SGBM- or RAFT-derived geometry.}
    \label{fig:printable_hybrid}
\end{figure}

The final printable mesh stage converted the reconstructed geometry into OBJ files for the first five SGBM-derived and the first five RAFT-derived Curiosity scenes for each completion strategy (Figure~\ref{fig:printable_hybrid}). By construction, the resulting meshes were single-component and watertight, with an average largest-component fraction of 1.0 (Table~\ref{tab:printable}).

\begin{table}[htbp]
    \centering
    \small
    \setlength{\tabcolsep}{4pt}
    \caption{Printable mesh statistics for Curiosity reconstructions. Scenes is the number of exported OBJ files. Vert. and Faces are average mesh vertices and faces. CC is the average number of connected components. LCC is the fraction of faces in the largest connected component.}
    \label{tab:printable}
    \begin{tabular}{lccccc}
        \toprule
        Method & Scenes & Vert. & Faces & CC & LCC \\
        \midrule
        Poisson & 10 & 185.7k & 374.4k & 1.0 & 1.0 \\
        Alpha & 10 & 92.1k & 186.1k & 1.0 & 1.0 \\
        Hybrid & 10 & 192.4k & 387.4k & 1.0 & 1.0 \\
        Diff.-fill & 10 & 95.8k & 193.2k & 1.0 & 1.0 \\
        \bottomrule
    \end{tabular}
\end{table}

\section{Discussion}

Our results show that benchmark performance on standard stereo datasets does not directly translate to reliable Martian terrain reconstruction. Although RAFT-Stereo performs better than SGBM on Middlebury, its Curiosity outputs show that dense disparities can still smooth terrain boundaries, increase photometric reprojection error, and shift depth statistics under planetary domain shift. This suggests that future rover stereo systems should evaluate edge alignment, uncertainty, and metric geometry in addition to standard disparity accuracy \cite{raftstereo, middlebury, sgm2008}. Classical methods such as SGBM remain useful because their failures are often sparse and easier to identify.

The occlusion estimation results show a tradeoff between local fidelity and global surface structure. Alpha shapes preserve more of the measured boundary structure but are sensitive to sampling density and can fragment, while Poisson reconstruction regularizes the point cloud into a smoother and more connected implicit surface \cite{edelsbrunner1994threedalpha, kazhdan2006poisson}. The deterministic diffusion-fill baseline provided a simple heuristic comparison, but its quality still depended strongly on the input depth. For scientific measurement, inferred surfaces should carry uncertainty or be separated from directly observed geometry.

\subsection{Limitations and failure cases}

The main limitation of this study is the lack of ground-truth 3D geometry for Curiosity terrain. Because ground-truth metric depth is unavailable, we rely on proxy metrics such as reprojection error, valid disparity ratio, edge alignment, and visible-surface consistency. These metrics are useful for diagnosing stereo behavior, but they cannot fully determine whether a reconstructed terrain patch is geometrically correct.

The pipeline also compounds errors across stages. Sparse or unstable disparities produce incomplete point clouds, which can cause alpha-shape reconstructions to fragment. Poisson reconstruction can fill gaps and produce more connected meshes, but this connectivity may come from smoothing or adding unsupported surfaces. The synthetic-ground-truth ablation demonstrated that this tradeoff persists even without stereo error: alpha shapes remained fragmented, while Poisson reconstruction stayed connected but less faithful to the known surface. Voxelization and marching cubes enforce watertight printable OBJ files, but they can hide upstream errors by incorporating uncertain reconstructions into the final object \cite{lorensen1987marching}. Future work should use synthetic Martian renderings, multi-view rover sequences, or physical scans to evaluate reconstructed meshes against known geometry.

\subsection{Ethical considerations}

This project has relatively low ethical risk, but it raises issues of transparency, misuse, and misattribution. Released models should label reconstructed regions as approximations. Methods for inferring hidden structure from partial imagery could also be repurposed for unintended uses such as surveillance. Because the input imagery comes from NASA public archives, any released models or printed artifacts should attribute the original data source and preserve visible documentation of the reconstruction process.

\section{Conclusion}

We evaluated an end-to-end pipeline for reconstructing and 3D printing Martian terrain models from NASA Curiosity stereo imagery. RAFT-Stereo performs better than SGBM on Middlebury and produces denser Curiosity disparities, but its dense predictions can smooth terrain boundaries and shift Curiosity depths toward shallower estimates under planetary domain shift. SGBM is less complete, but its surviving disparities align more closely with local image structure. For occlusion estimation, alpha shapes best preserve observed geometry while often fragmenting, and Poisson reconstruction reliably produces connected surfaces by sacrificing geometric fidelity. A practical outcome of the pipeline is printability: voxel repair and marching-cubes extraction successfully converted the evaluated subset of completion outputs into watertight OBJ files suitable for physical fabrication. These printed models should be treated as interpretable approximations, not metric reconstructions. Accurate Martian terrain reconstruction will require domain-specific validation that tests stereo, completion, and mesh repair against known 3D geometry.

\section{Code availability}

All code for this project is available at \url{https://github.com/josiexw/6s058}. The dataset and output image/OBJ files are available at \url{https://drive.google.com/file/d/1rScqlOSbZqFMa16_9_kRi62OMqAFD9cl/view?usp=sharing}.

{
    \small
    \bibliographystyle{ieeenat_fullname.bst}
    \bibliography{main}
}

\setcounter{page}{1}
\maketitlesupplementary

\section{Metric Definitions}
\label{app}

\paragraph{Stereo metrics.}
For Curiosity images, the valid disparity fraction and valid depth fraction were defined as
\begin{align}
    r_d &= \frac{|\Omega_d|}{HW}, &
    r_z &= \frac{|\Omega_z|}{HW}.
\end{align}
The corresponding valid sets were
\begin{align}
    \Omega_d
    &= \{(u,v): d(u,v)>0,\ d(u,v)\ \mathrm{finite}\}, \\
    \Omega_z
    &= \{(u,v): z_{\min}<z(u,v)<z_{\max},\ z(u,v)\ \mathrm{finite}\}.
\end{align}
In these expressions, $H$ and $W$ denote the image height and width. The reprojection valid fraction was defined as
\begin{align}
    R_{\mathrm{reproj}} &=
    \frac{|\Omega_{\mathrm{reproj}}|}{HW}, \\
    \Omega_{\mathrm{reproj}} &=
    \{(u,v): d(u,v)>0,\ u-d(u,v)\in[0,W-1]\}.
\end{align}
Over the valid reprojection domain $\Omega$, the photometric reprojection error was
\begin{equation}
    E_{\mathrm{photo}} =
    \frac{1}{|\Omega|}
    \sum_{(u,v)\in\Omega}
    |I_L(u,v)-I_R(u-d(u,v),v)|.
\end{equation}
The disparity-gradient statistic and edge-alignment score were
\begin{align}
    S_d &=
    \mathrm{median}_{(u,v)\in\Omega}
    \left(|\partial_x d(u,v)|+|\partial_y d(u,v)|\right), \\
    A_{\mathrm{edge}} &=
    \mathrm{corr}\left(\|\nabla I_L\|,\|\nabla d\|\right).
\end{align}
We also reported the median valid disparity $\tilde{d}$ and median valid depth $\tilde{z}$ over the corresponding valid pixels.

\paragraph{Middlebury benchmark metrics.}
For Middlebury, the valid prediction ratio was defined as
\begin{align}
    R_{\mathrm{pred}} &=
    \frac{|\Omega_{\mathrm{pred}}|}{|\Omega_{\mathrm{gt}}|}, \\
    \Omega_{\mathrm{pred}} &=
    \{(u,v)\in\Omega_{\mathrm{gt}}:
    d(u,v)>0,\ d(u,v)\ \mathrm{finite}\}.
\end{align}
In this setting, $\Omega_{\mathrm{gt}}$ denotes the set of valid ground-truth disparity pixels. Disparity error was evaluated using
\begin{align}
    \mathrm{MAE} &=
    \frac{1}{|\Omega_{\mathrm{gt}}|}
    \sum_{(u,v)\in\Omega_{\mathrm{gt}}}
    |d(u,v)-d^*(u,v)|, \\
    \mathrm{RMSE} &=
    \sqrt{
    \frac{1}{|\Omega_{\mathrm{gt}}|}
    \sum_{(u,v)\in\Omega_{\mathrm{gt}}}
    (d(u,v)-d^*(u,v))^2
    }.
\end{align}
The bad-pixel rate at threshold $\tau$ was defined as
\begin{align}
    \mathrm{Bad}_{\tau} &=
    \frac{|\Omega_{\mathrm{bad},\tau}|}{|\Omega_{\mathrm{gt}}|}, \\
    \Omega_{\mathrm{bad},\tau} &=
    \{(u,v)\in\Omega_{\mathrm{gt}}:
    |d(u,v)-d^*(u,v)|>\tau\}.
\end{align}

\paragraph{Geometry completion metrics.}
Let $P$ denote the reference point set and $Q$ the point set sampled from the reconstructed mesh. The one-way nearest-neighbor distance was
\begin{equation}
    D_{P\rightarrow Q} =
    \frac{1}{|P|}
    \sum_{p\in P}
    \min_{q\in Q}\|p-q\|_2.
\end{equation}
The Chamfer distance and normalized Chamfer distance were defined as
\begin{align}
    D_{\mathrm{Chamfer}}(P,Q)
    &= D_{P\rightarrow Q}+D_{Q\rightarrow P}, \\
    D_{\mathrm{norm}}(P,Q)
    &= \frac{D_{\mathrm{Chamfer}}(P,Q)}{\mathrm{diag}(\mathcal{B})},
\end{align}
where $\mathrm{diag}(\mathcal{B})$ denotes the bounding-box diagonal of the reference point set. Visible coverage and mesh novelty at threshold $\epsilon$ were defined as
\begin{align}
    C_{\epsilon} &=
    \frac{|\{p\in P: \min_{q\in Q}\|p-q\|_2 < \epsilon\}|}{|P|}, \\
    N_{\epsilon} &=
    \frac{|\{q\in Q: \min_{p\in P}\|q-p\|_2 > \epsilon\}|}{|Q|}.
\end{align}

\paragraph{Mesh topology and printability metrics.}
For mesh topology and printability, we reported vertex count, triangle or face count, connected components, surface area, and the largest-component fraction. The largest-component fraction was defined as
\begin{equation}
    R_{\mathrm{LCC}} =
    \frac{|F_{\mathrm{largest}}|}{|F_{\mathrm{all}}|},
\end{equation}
where $F_{\mathrm{largest}}$ is the set of faces in the largest connected component and $F_{\mathrm{all}}$ is the set of all mesh faces. Watertightness was reported as a binary indicator of whether the exported printable mesh was watertight.

\end{document}